\documentclass[journal]{IEEEtran}
\usepackage{longfigure}
\usepackage{flushend}
\usepackage{forest}
\usepackage{diagbox}
\usepackage{slashbox}
\usepackage{threeparttable}
\usepackage{fontawesome}
\usepackage{multirow}
\usepackage{fullpage}
\usepackage{bbm}
\usepackage{enumitem}
\usepackage{hyperref}
\usepackage{amsthm}
\usepackage[square, comma, sort&compress, numbers]{natbib}
\usepackage{amsmath,amssymb,amsfonts}
\usepackage{algorithmic}
\usepackage{graphicx}
\usepackage{textcomp}
\usepackage{subfigure}
\usepackage{float}
\usepackage{xcolor}
\usepackage[centerlast]{caption2}
\newtheorem*{remark}{Remark}
\usepackage{graphicx}

\usepackage{boxhandler}

\ifCLASSINFOpdf
\else
\fi
\begin{document}
%
\title{Classification using Hyperdimensional Computing: A Review}

%
%
%

\author{Lulu~Ge  
        and Keshab K. Parhi
\thanks{L. Ge and K.K. Parhi are with the Department
of Electrical and Computer Engineering, University of Minnesota,
Minneapolis, MN 55455, USA e-mail: \{ge000567, parhi\}@umn.edu}}
\maketitle

\begin{abstract}
Hyperdimensional (HD) computing is built upon its unique \textcolor{black}{data type} \textcolor{black}{referred to as} \textit{hypervectors}. The dimension of these hypervectors is typically in the range of tens of thousands. Proposed to solve cognitive tasks, HD computing aims at calculating similarity among its data. Data transformation is realized by three operations, including addition, multiplication and permutation. Its ultra-wide data representation introduces redundancy against noise. Since information is evenly distributed over every bit of the hypervectors, HD computing is inherently robust. Additionally, due to the nature of those three operations, HD computing leads to fast learning ability, high energy efficiency and acceptable accuracy in learning and classification tasks. This paper introduces the background of HD computing, and reviews the data representation, data transformation, and similarity measurement. The orthogonality in high dimensions presents opportunities for flexible computing. To balance the tradeoff between accuracy and efficiency, strategies include but are not limited to encoding, retraining, binarization and hardware acceleration. Evaluations indicate that HD computing shows great potential in addressing problems using data in the form of letters, signals and images. HD computing especially shows significant promise to replace machine learning algorithms as a light-weight classifier in the field of internet of things (IoTs).
\end{abstract} 


%
\IEEEpeerreviewmaketitle

\section{Introduction}\label{s:1}
%
%
%
%
\textcolor{black} {\IEEEPARstart {T}{he}} hyperdimensional (HD) emergence of hyperdimensional (HD) computing is
based on the cognitive model developed by Kanerva \cite{kanerva1988sparse}. HD computing grew out of cognitive science in answer to the binding problem of connectionist (neural-net) models. When variables and their values are superposed over the same vector, representing which value is associated with which variable requires a formal model. This was initially solved using tensor product variable binding by Smolensky \cite{smolensky1990tensor} and later by Plate\cite{plate1995holographic} using holographic reduced representation (HRR). The advantage of HRR over tensor product is that it keeps vector dimensionality constant. Systems based on these representations go by many names: HRR, HD, binary spatter code (BSD) \cite{kanerva1997fully}, binary sparse distributed code (BSDC) \cite{rachkovskij2001binding}, multiply-add-permute (MAP) \cite{gayler1998multiplicative}, vector symbolic architecture (VSA) \cite{schlegel2020comparison}, and semantic pointer architecture. All rely on high dimensionality, randomness, abundance of nearly orthogonal vectors and computing in superposition.

Instead of computing traditional numerical values, HD computing performs cognition tasks---such as face detection, language classification, speech recognition, image classification, etc---by representing different types of data using hypervectors, whose dimensionality is in the thousands, e.g., 10,000-$d$, \textcolor{black}{where $d$ refers to dimensionality}. \textcolor{black}{The human brain contains about 100 billion neurons and 1000 trillion synapses; therefore all possible {\em states} of a human brain can be described by a high-dimensional vector. In that sense, HD computing is a form of brain-inspired computing.} Randomly or pseudo-randomly defined, these \textcolor{black}{hypervectors} are composed of independent and identically distributed ($i.i.d.$) components, which can be binary, integer, real or complex \cite{hersche2018exploring}. As a brain-inspired computing model, HD computing is robust, scalable, energy efficient and requires less time for training and inference \textcolor{black}{\cite{rahimi2017high}}. These features are a result of its ultra-wide data representation and underlying mathematical operations. One thing that should be emphasized is the concept of {\em orthogonality} of the hypervectors.

The remainder of this paper is organized as follows. \textcolor{black}{Section \ref{s:2}} presents the background on HD computing, including the data representation, data transformation and similarity measurement. Section \ref{s:3} illustrates the general methodology in HD computing and its applicability in learning and inference tasks. Then two common encoding methods to form hypervectors from the input data are presented, and strategies to improve accuracy and/or efficiency are pointed out. \textcolor{black}{Some classical applications as well as several sophisticated designs are reviewed in Section \ref{s:4}. Possible future directions of HD computing are also pointed out in this section.} Finally, Section \ref{s:5} concludes the paper.

\begin{table*}[htbp]
\centering
\caption{Comparisons between classical computing and HD computing \textcolor{black}{for Classification.}}
\begin{tabular}{c||c|c}
\hline
\textbf{Computing Paradigm} & \textbf{Classical Computing} & \textbf{HD Computing }\\
\hline
\hline
\textbf{Data Type} & Bit & Hypervector \\
\hline
\textbf{Data Transformation} & Addition, Multiplication, Logic & \textcolor{black}{Add-Multiply-Permute} \\
\hline
\textbf{\textcolor{black}{Storage}} & \textcolor{black}{Memory} & Item Memory, Associative Memory \\
\hline
\textbf{\textcolor{black}{Training}} & \textcolor{black}{Weights} & Class Hypervectors \\
\hline
\textbf{\textcolor{black}{Testing}} & \textcolor{black}{Run Pre-trained Classifier} & Associate Query Hypervectors with Class Hypervectors \\
\hline
\textbf{\textcolor{black}{Model Complexity}} & \textcolor{black}{High} & Low \\
\hline
\textbf{\textcolor{black}{Accuracy}} & \textcolor{black}{Very High} & Acceptable \\
\hline
\textbf{\textcolor{black}{Feature Encoding}} & \textcolor{black}{Easy} & Difficult \\
\hline
\textbf{ Number of Features}  & Many & One \\
\hline
\end{tabular}\label{t:com}
\end{table*}

\section{Background on HD Computing} \label{s:2}
In this section, we review HD computing and present a comparison between HD and classical computing. We also describe the similarity metrics for hypervectors and typical mathematical operations used in HD computing.
\vspace{-8pt}
\subsection{Classical Computing vs HD Computing}
Data representation, data transformation and data retrieval play an important role in any computing system. To be more specific, classical computing deals with bits. Each bit is 0 or 1. This can be realized by the absence or presence of electric charge. \textcolor{black}{In terms of computation, data transformation is inevitable}. The arithmetic/logic unit (ALU) computes new data using logical operation and four arithmetic operations, including addition, subtraction, multiplication and division \cite{bryant2015computer}. The main memory allows the data to be written and read. Compared to classical computing, HD computing employs hypervectors as its data type, \textcolor{black}{whose dimensionality is typically in the thousands}. These ultra-wide words introduce redundancy against noise, and are, therefore, inherently robust. 

To transform data, HD computing performs three operations: multiplication, addition and permutation. HD computing transforms the input hypervectors, which are pre-stored in the item memory to form associations or connections. In a classification problem, the hypervectors associated with classes are trained during training process. During the testing process, the test hypervectors are compared with the class hypervectors. The hypervectors generated from training data are referred to as  {\em class} hypervectors and are stored in the associative memory, while those generated from the test data are \textcolor{black}{referred to as} {\em query} hypervectors. \textcolor{black}{An associative search is performed to make a prediction as to which class a given query hypervector most likely belongs.} A comparison between the classical and HD computing paradigms is summarized in Table \ref{t:com}. Traditional classification methods achieve high accuracy using complex models. Training these models typically takes longer time and requires more energy consumption. The models in HD classification are simpler and can be trained in less time with high energy efficiency. However, their accuracy is acceptable, though not as high as traditional models. This is because the accuracy is dependent on feature encoding which is not as well understood as traditional classification. 

\subsection{Data Representation}
Data points of HD computing correspond to hypervectors---vectors of bits, integers, real or complex numbers. These are roughly divided into two categories: binary and non-binary. For non-binary hypervectors, bipolar and integer hypervectors are more commonly employed. Generally speaking, non-binary HD algorithms achieve higher accuracy, while the binary counterpart is more hardware-friendly and has higher efficiency (see also \cite{patyk2011comparison}).

\subsection{Similarity Measurement}

As shown in Table \ref{t:sim}, two common similarity measurements are adopted in the existing HD algorithms, namely, cosine similarity and Hamming distance. Other similarity measures include dot product (e.g., in MAP) and overlap (e.g., in BSDC).
\begin{table}[htbp]
  \centering
  \caption{Similarity Measurements in HD Computing.}
\begin{tabular}{c||c|c}
\hline
\textbf{Measurement} & \textbf{Similar} & \textbf{Orthogonal} \\
\hline
\hline
\textbf{Hamming distance} & 0 & 0.5\\
\hline
\textbf{Cosine similarity} & 1 & 0\\
\hline
\end{tabular}\label{t:sim}
\end{table}

For non-binary hypervectors, {\em cosine similarity}, defined by Eq. (\ref{eq0}), is used to measure their similarity, focusing on the angle and ignoring the impact of the magnitude of hypervectors, where $|\cdot|$ denotes the magnitude. Unlike the inner product operation \cite{olver2018applied} of two vectors that affects magnitude and orientation, the {\em cosine similarity} only depends on the orientation. In most HD algorithms with non-binary hypervectors, cosine similarity is more often used than inner product. Moreover, \textcolor{black}{when $\text{cos}(A, B)$ is close to $1$, this implies an extremely high level of similarity. For example, $\text{cos}(A, B)=1$ indicates two hypervectors $A$ and $B$ are identical}. When they are at right angle, then $\text{cos}(A, B)=0$, and the two orthogonal vectors are considered dissimilar.
\begin{equation}\label{eq0}
    \text{cos}(A,B)=\frac{A \cdot B}{|A||B|} 
\end{equation}

For binary hypervectors with dimensionality $d$, whose components are \textcolor{black}{either $0$ or $1$}, normalized Hamming distance calculated in Eq. (\ref{eq1}) is used to measure their similarity \cite{hersche2018exploring}. When the Hamming distance of two hypervectors is close to $0$, then they are defined as similar. For example, $\text{Ham}(A,B)=0$ indicates every single bit is same at each position, and $A$ and $B$ are \textcolor{black}{identical}. When $\text{Ham}(A,B)=0.5$, $A$ and $B$ are orthogonal or dissimilar. $\text{Ham}(A,B)=1$ when $A$ and $B$ are diametrically opposed. 
\vspace{-8pt}
\begin{equation}\label{eq1}
    \text{Ham}(A,B)=\frac{1}{d}\sum_{i=1}^{d}1_{A(i) \neq B(i)}
\end{equation}
\vspace{-12pt}
\begin{figure}[H]
\centerline{\includegraphics[width=0.74\linewidth]{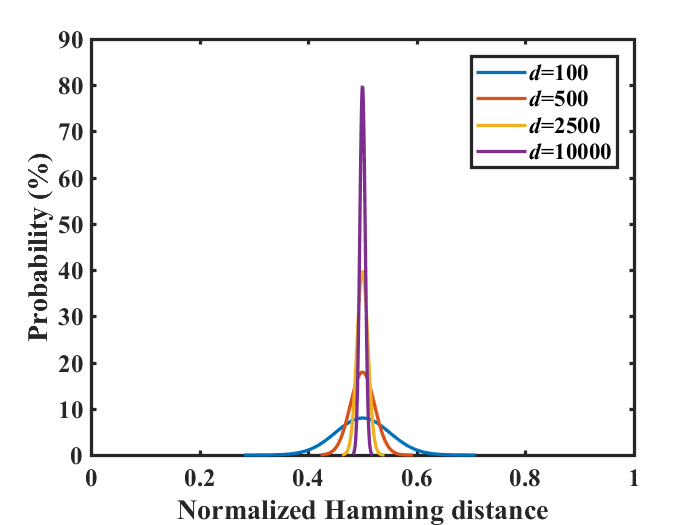}}
\caption{Orthogonality in high dimensions \cite{kanerva1988sparse,datta2019programmable,widdows2015reasoning}}
\label{f0}
\end{figure} 


One thing that should be emphasized is orthogonality in high dimensions. To put it simply, the randomly generated hypervectors are nearly orthogonal to each other when the dimensionality is in the thousands. Take binary hypervectors as an example. Assume $X$ and $Y$ are chosen independently and uniformly from $\{0,1\}^d$ and the probability $p$ of any bit being 1 is 0.5. Then $\text{Ham}(X,Y)$ is binomially distributed. Fig. \ref{f0} shows the probability density function (PDF) of $\text{Ham}(X,Y)$ for 15,000 pairs of randomly selected binary vectors with different dimensions $d$. As $d$ increases, more vectors become orthogonal. Such orthogonality property is of great interest because orthogonal hypervectors are dissimilar. Moreover, operations performed on these orthogonal hypervectors can form associations or relations. 
\vspace{-10pt}

\subsection{Data Transformation}
\vspace{-4pt}
Three types of operations, add-multiply-permute, are employed in HD computing. The inverse operation for multiplication is also referred to as {\em release} \cite{widdows2015reasoning}. The release operation is also used to denote inverse addition. Each operation processes and generates $d$-\textcolor{black}{dimensional} hypervectors. \textcolor{black}{In the following, we illustrate examples of data transformations using binary hypervectors. Without doubt, data transformation can also be employed to non-binary hypervectors, which is in essence similar to the manipulations over binary hypervectors. The only difference is from the point of hardware; for binary hypervectors, the pointwise multiplication can be realized by an exclusive or (\textsc{xor}) gate.} 

\begin{figure*}[htpb]
\centering
\subfigure[$A$+$B$+$C$=$X$.]{
\includegraphics[width=0.31 \linewidth]{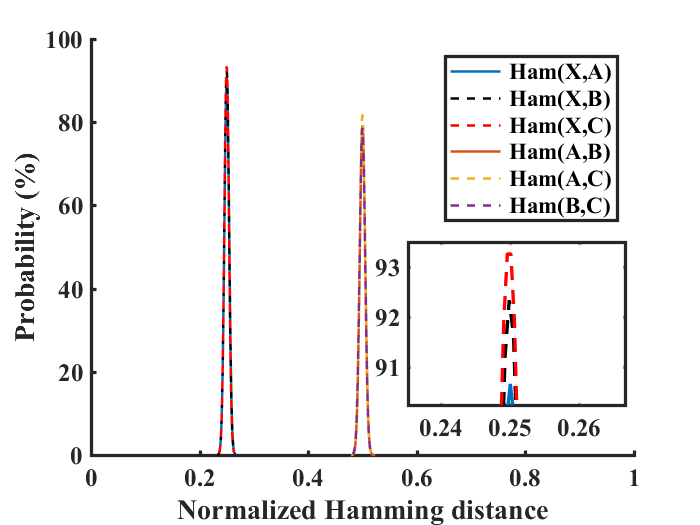}}
    \subfigure[$A$+$B$=$X$ in favor of 0.]{\label{f:pa:2}
\includegraphics[width=0.31 \linewidth]{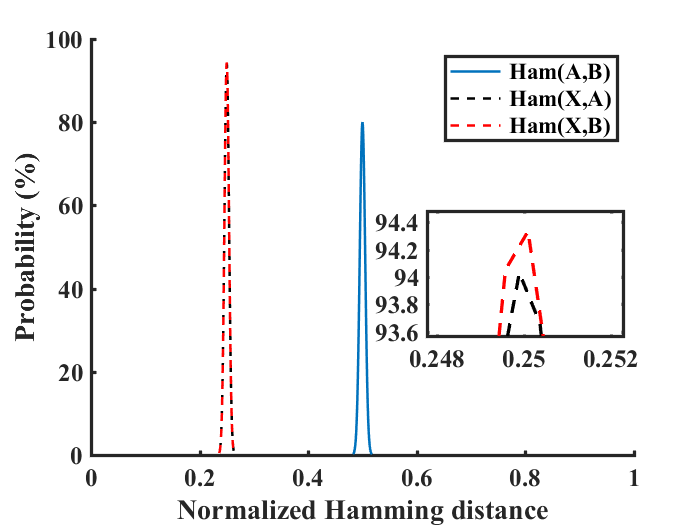}}
    \subfigure[$A$+$B$=$X$ in favor of 1.]{\label{f:pa:3}
\includegraphics[width=0.31 \linewidth]{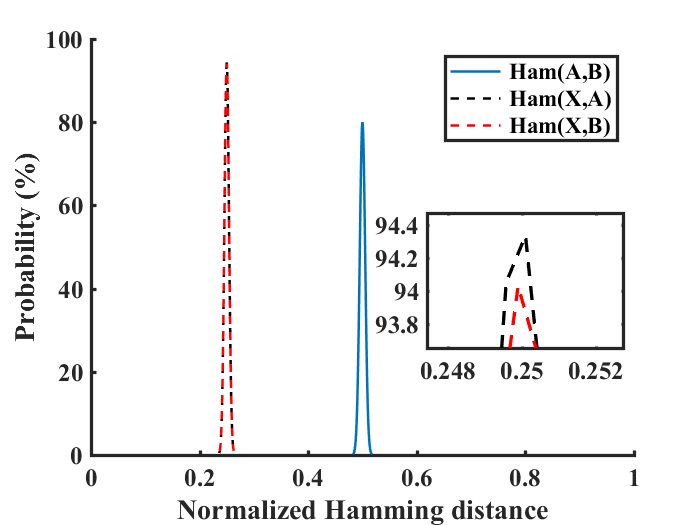}}
\vspace{-12pt}
\caption{\textcolor{black}{Hamming distance distribution of addition for 10,000-bit hypervectors over 3000 cases. (a) Addition over odd number of hypervectors; (b) and (c) shows the addition over even number favoring 0 and 1, respectively.}}\label{f:pa}
\captionStyle{n}{l}
\end{figure*}

\subsubsection{Addition}
Pointwise addition, also \textcolor{black}{referred to as} \textit{bundling}, computes a hypervector $Z$ using Eq. (\ref{eq2}) from the input hypervectors $\{X_1,X_2,\cdots, X_n\}$. Compared to random hypervectors, the generated $Z$ is maximally similar to the $n$ inputs  $X_1,X_2,\cdots, X_n$, i.e., Hamming distance between $Z$ and any of the $n$ inputs is at a minimum. 
\vspace{-3pt}
\begin{equation}\label{eq2}
 \begin{aligned}
Z=[X_1+X_2+\cdots+X_n],
\end{aligned}
\vspace{-3pt}
\end{equation}
where $[ \cdot ]$ indicates the sum hypervector $Z$ is \textcolor{black}{thresholded and binarized} to $\{0,1\}^d$ based on the {\em majority rule}. For convenience, Eq. (\ref{eq:add}) shows an example for the pointwise addition of three 10-bit binary vectors.
\begin{equation}\label{eq:add}
 \begin{aligned}
A=0 \;	 0 \;	 0 \;	0 \;	1 \;	1 \;	0 \;	0 \;	1 \;	1,\\
B=1	 \; 0 \;	1 \;	1 \;	0 \;	0 \;	0 \;	1 \;	0 \;	1,\\
C=0	 \; 0 \;	1 \;	0 \;	1 \;	0 \;	1 \;	1 \;	0 \;	1,\\
\hline
[A+B+C]=0  \; 0  \;1 \; 0  \;1  \;0  \;0  \;1 \; 0 \; 1.
\end{aligned}
\end{equation}


Generally speaking, the addition over odd number of hypervectors has no ambiguity, whereas the addition over an even number can favor either $0$ or $1$ using the majority function defined in Eq. (\ref{eq:majority}). However, this approach may lead to a biased result for adding two hypervectors. Therefore, the bias in adding even number of hypervectors is usually reduced by adding an extra random vector \cite{schmuck2019hardware}. Fig. \ref{f:pa}  illustrates addition of 10,000-dimensional random hypervectors repeated for 3,000 times. Comparing Fig. \ref{f:pa:2} to Fig. \ref{f:pa:3}, we see that specifying in favor of 0 or 1 has little impact over addition. It can be observed from Fig. \ref{f:pa} that the sum is nearly equally similar to the input operands.
\vspace{-4pt}
\begin{equation}\label{eq:majority}
\resizebox{.9\hsize}{!}{$
 \text{Majority}(p_1,\cdots,p_n)=
 \left\{
     \begin{array}{lr}
      \lfloor \frac{1}{2}+\frac{(\sum_{i=1}^n p_i)-\frac{1}{2}}{n} \rfloor, \;\text{favor 0},\\
     \lfloor  \frac{1}{2}+\frac{\sum_{i=1}^n p_i}{n} \rfloor, \; \text{favor 1}. \\
    \end{array}
\right.$}
\end{equation}
\subsubsection{Multiplication}
\begin{figure}[H]
\centerline{\includegraphics[width=0.74\linewidth]{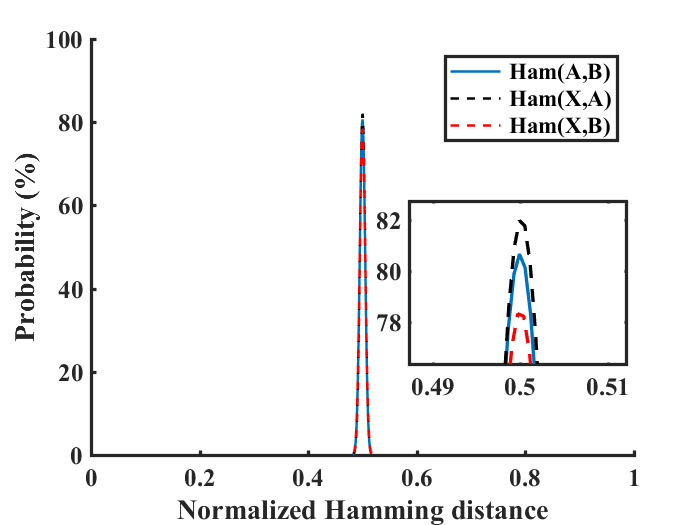}}
\vspace{-3pt}
\caption{\textcolor{black}{Hamming distance distribution of multiplication $X=A \oplus B$ for 10,000-bit hypervectors over 3000 cases.}}\label{f:pmu}
\end{figure}

Pointwise multiplication, also called \textit{binding}, aims to form associations between two related hypervectors. $A$ and $B$ are bound together to form $X=A \oplus B$, which is approximately orthogonal to both $A$ and $B$, \textcolor{black}{where $\oplus$ represents the $\textsc{xor}$ operation}. Eq. (\ref{eq:mult}) shows the pointwise multiplication of two 10-bit binary vectors. In a more general case, as shown in Fig. \ref{f:pmu}, for two randomly generated 10,000-bit binary hypervectors, their pointwise multiplication result is dissimilar to both of them.
\vspace{-3pt}
\begin{equation}\label{eq:mult}
 \begin{aligned}
A=0	 \; 0 \;	0 \;	0 \;	1	 \;1 \;	0 \;	0 \;	1 \;	1,\\
B=1	 \; 0 \;	1 \;	1 \;	0 \;	0 \;	0 \;	1 \;	0 \;	1,\\
\hline
A \oplus B =1 \;	 0	 \;1 \;	1	 \;1 \;	1 \;	0 \;	1 \;	1	 \;0.\\
\end{aligned}
\end{equation}
\vspace{-12pt}

\subsubsection{Permutation}
\begin{figure}[H]
\centering
\includegraphics[width=0.74\linewidth]{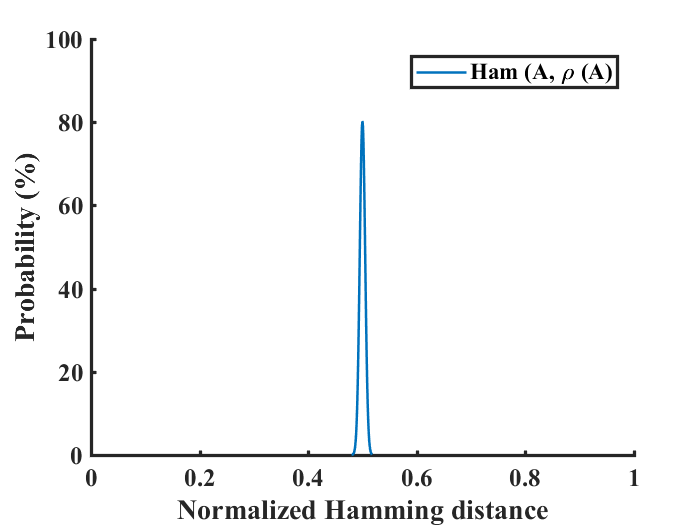}
\caption{\textcolor{black}{Hamming distance distribution of permutation for 10,000-bit hypervectors over 3000 cases.}} \label{f:pp}
\end{figure}

Permutation $\rho$ is a unique unary operation for HD computing, which shuffles the hypervector, let us say $A$. The resulting permuted hypervector $\rho(A)$ is \textcolor{black}{quasi-orthogonal to the initial $A$, i.e, the normalized Hamming distance is close to $0.5$}. Mathematically, permutation can be realized by multiplying a permutation matrix. As a specific permutation, circular shift is widely employed for its friendly hardware implementation. Eq. (\ref{eq:p}) shows a circular shift of a 10-bit binary vector \textcolor{black}{with Ham$(A,\rho(A))=0.4$. Expected Hamming distance is supposed to be 0.5 for ultra-wide hypervectors.} Fig. \ref{f:pp} indicates the permutation result shows dissimilarity with the original 10,000-bit hypervector.
\begin{equation}\label{eq:p}
 \begin{aligned}
A=0	 \; 0 \;	0 \;	0 \;	1	 \;1 \;	0 \;	0 \;	1 \;	1,\\
\hline
\rho(A) =	1\;0	 \; 0 \;	0 \;	0 \;	1	 \;1 \;	0 \;	0 \;	1 .\\
\end{aligned}
\end{equation}
\vspace{-8pt}
\textbf{Examples.} We illustrate applications of above operations. For more details, please refer to \cite{kanerva2009hyperdimensional}. Assume that $A,B,C,P,S,X,Y,Z$ represent 10,000-$d$ random hypervectors:  

\begin{itemize}
    \item {\em Encode} a pair: \textcolor{black}{To encode ``$x=a$'', where $x$ is a variable with numerical value same as $a$, use multiplication to bind their corresponding hypervectors $X$ and $A$.} The encoding is represented by the generated hypervector $P=X \oplus A$.
\item {{\em \textcolor{black}{Release}} the value from the pair:} 
\begin{equation}\label{eq3}
 \begin{aligned}
 X \oplus P=\underbrace{X \oplus( X \oplus A)}_{X \oplus  X \;\text{cancels out}}=A
\end{aligned}
\end{equation}
\item {\em \textcolor{black}{Represent}} a set:
Given the set $s=\{a,b,c\}$, we have
\begin{equation}\label{eq4}
 \begin{aligned}
 S=[A+B+C]
\end{aligned}
\end{equation}
\item {{\em Encode} a data record:} Given a record with a set of bound pairs $d=$`$(x=a)\&(y=b)\&(z=c)$', the record is encoded as:
\begin{equation}\label{eq5}
 \begin{aligned}
 D=[X\oplus A+Y \oplus B+Z \oplus C]
\end{aligned}
\end{equation}
\item {{\em Extract} the value from a record}: To retrieve the value of $x$:
\begin{equation}\label{eq6}
 \begin{aligned}
 A'&=X \oplus D \\
 &= \underbrace{X  \oplus [X\oplus A+Y \oplus B+Z \oplus C]}_{\text{distributed}}\\
 &= X  \oplus X\oplus A+ X  \oplus Y \oplus B+ X  \oplus Z \oplus C\\
 &= \underbrace{X  \oplus X\oplus A}_{=A}+ \underbrace{\big( X  \oplus Y \oplus B+ X  \oplus Z \oplus C \big)} _{\text{noise}}\\
 & \approx  A\\
\end{aligned}
\end{equation}
\item {{\em Encode} a sequence}: Given $(a,b)$, then 
\begin{equation}\label{eq7}
 \begin{aligned}
AB=\rho(A) \oplus B
\end{aligned}
\end{equation}
\item {{\em Extend} the sequence}: Extend $(a,b)$ to $(a,b,c)$ using:
\begin{equation}\label{eq8}
 \begin{aligned}
 ABC &=\rho(AB) \oplus C\\
 &=\rho\big(\rho(A)\big)\oplus  \rho(B) \oplus C
\end{aligned}
\end{equation}
\item {{\em Extract} the first element of the sequence}:
\begin{equation}\label{eq9}
 \begin{aligned}
 & \rho^{-1}\rho^{-1}(ABC \oplus BC) \\
 &= \rho^{-1}\rho^{-1}(\rho\big(\rho(A)\big) \oplus  \rho(B) \oplus C \oplus \rho(B) \oplus C)\\
 &= \rho^{-1}\rho^{-1}(\rho\big(\rho(A)\big)\\
 &=A
\end{aligned}
\end{equation}
where \textcolor{black}{$\rho^{-1}$} is the inverse operation of permutation $\rho$. 
\end{itemize}
\section{Learning and Classification \textcolor{black}{By HD Computing}}\label{s:3}
The first wave of using HD for classification started in 1990s  \cite{rachkovskij2007linear,kussul1998application,kussul1991image,rachkovskij1990audio}. The current applications of HD for classification can be interpreted as the second wave.

\subsection{The HD Classification Methodology}
\begin{figure}[ht] 
\centerline{\includegraphics[width=\linewidth]{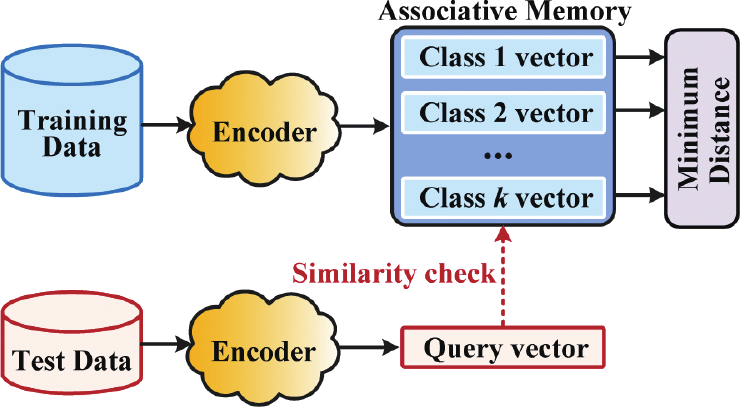}}
\caption{Classification overview with HD computing \cite{rahimi2016robust}}.
\label{f01}
\end{figure}

A system diagram for the classification tasks using HD computing is shown in Fig. \ref{f01}. In general, \textcolor{blue}{\textbf{\textit{1).}}} during the learning phase, the encoder employs randomly generated hypervectors (pre-stored in the item memory) to map the training data into HD space. A total of $k$ class hypervectors are trained and stored in the associative memory. \textcolor{blue}{\textbf{\textit{2).}}} During the inference phase, the encoder generates the query hypervector for each test data. Then the similarity check is conducted in the associative memory between the query hypervector and every pre-trained class hypervector. Finally, the label with the closest distance is returned.

\subsection{Encoding Methods for HD Computing}
HD computing can address various types of input data, including letters, signals and images. However, we need to map those input data into hypervectors, and this process corresponds to encoding. The encoding process is somewhat similar to extraction of features. Among the existing HD algorithms, the two encoding methods commonly used include {\em record-based encoding} and {\em $N$-gram-based encoding}. A toy example related speech signals is used for illustration.

Using Mel-frequency cepstral coefficients (MFCCs) \cite{logan2000mel}, the voice information stored in continuous signals can be mapped into the frequency domain. A feature vector with $N$ elements can be obtained. Each element has its feature value, which is evenly discretized or quantized from $\{F_{min}, F_{max}\}$ to $m$ different levels. 

\begin{figure}[ht] 
\centerline{\includegraphics[width=0.85\linewidth]{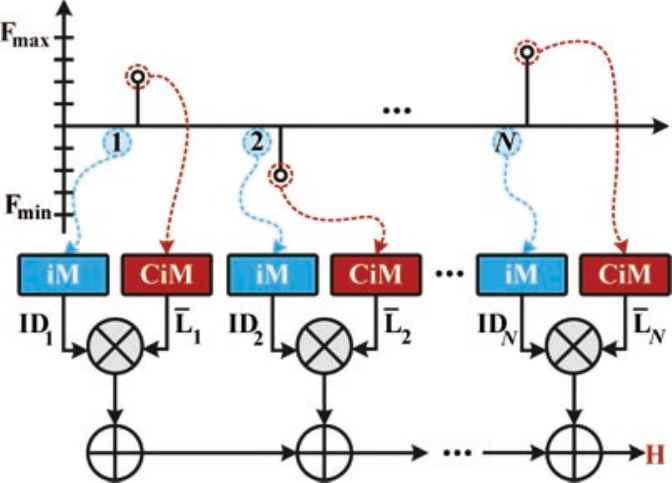}}
\caption{Record-based encoding \cite{rahimi2016hyperdimensional}. Note iM refers to item memory, which stores the position hypervectors, and CiM refers to continuous item memory \cite{datta2019programmable}, which stores level hypervectors.}
\label{f:record}
\end{figure}

\subsubsection{Record-based Encoding} This encoding method employs two types of hypervectors, representing the feature position and feature value, respectively. It may be noted that a variation of record-based encoding based on permutations and a chain of binding operations was proposed in \cite{kleyko2014brain}. In this encoding, \textbf{position hypervectors} $\text{ID}_{i}$ are randomly generated to encode the feature position information in a feature vector, where $1 \leq i \leq N$. The feature value information is quantized to $m$ \textbf{level hypervectors} $\{\mathbf{L}_1,\mathbf{L}_2,\cdots, \mathbf{L}_m \}$. \textcolor{black}{For an $N$-dimensional feature, a total of $N$ level hypervectors $\mathbf{\bar{L}}_i$ should be generated, which are chosen from $m$ level hypervectors $\{\mathbf{L}_1,\mathbf{L}_2,\cdots, \mathbf{L}_m \}$ based on the feature value.} Note that, position hypervectors $\text{ID}_{i}$ are orthogonal to each other, while level hypervectors $\{\mathbf{L}_1,\mathbf{L}_2,\cdots, \mathbf{L}_m \}$ are supposed to have correlations between the neighbours. To realize this, in \cite{imani2019quanthd} the first level hypervector $\mathbf{L}_1$ represents the feature value $F_{min}$. Then each time, $d/m$ randomly selected bits are flipped to generate the next level hypervector, where $d$ is the dimensionality of the hypervectors. \textcolor{black}{The continuous bit-flipping was first introduced in \cite{rahimi2016hyperdimensional} and later followed by other use cases \cite{rahimi2017hyperdimensional2,moin2018emg, imani2017voicehd}.} This bit-flipping approach ensures the correlations between neighbor levels, while the last level hypervector $\mathbf{L}_m$ is nearly orthogonal to $\mathbf{L}_1$. The encoding occurs by binding each position hypervector with its level hypervector. As described in Eq. (\ref{eq:rec}), the final encoding hypervector $\mathbf{H}$ can be obtained by adding these results together. The entire encoding process is illustrated in Fig. \ref{f:record}.

\begin{equation}\label{eq:rec}
 \begin{aligned}
& \mathbf{H} = \mathbf{\bar{L}}_1 \oplus \text{ID}_1 + \mathbf{\bar{L}}_2 \oplus \text{ID}_2 + \cdots + \mathbf{\bar{L}}_N \oplus \text{ID}_N,\\
& \mathbf{\bar{L}}_i  \in \{\mathbf{L}_1,\mathbf{L}_2,\cdots, \mathbf{L}_m \}, \; \text{where} \; 1 \leq i \leq N.
\end{aligned}
\end{equation}

\begin{figure}[ht] 
\centerline{\includegraphics[width=0.9\linewidth]{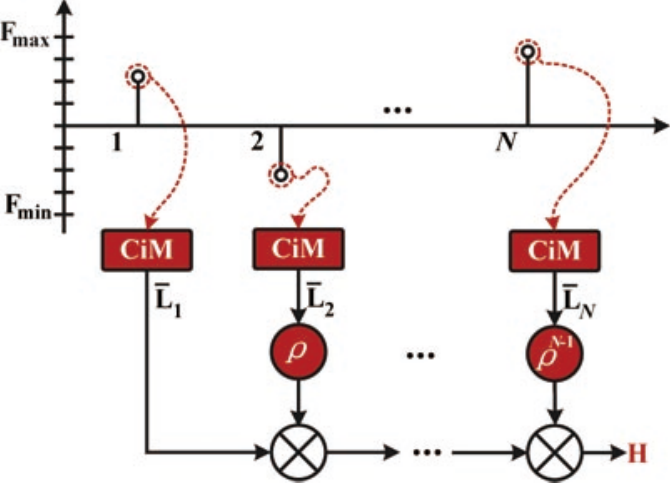}}
\caption{\textcolor{black}{$N$-gram-based} encoding \cite{imani2018hierarchical}. CiM stores level hypervectors which are mutually orthognal.}
\label{f:ngram}
\end{figure}

\begin{figure*}
\flushleft
\begin{forest} 
for tree={
    edge path={\noexpand\path[\forestoption{edge}] (\forestOve{\forestove{@parent}}{name}.parent anchor) -- +(0,-12pt)-| (\forestove{name}.child anchor)\forestoption{edge label};}
}
[HD Computing
  [Accuracy
    [Encoding ] 
    [Retraining]
    [Non-binary]
    ]
[Efficiency
  [Algorithm
    [Binarization]
    [Quantization]
    [Sparsity]]
  [Hardware
    [In-memory]
    [Nano Tech
    [CNFET] [RRAM] [3D intergration]
    ]
    [FPGA]
         ]]]
\end{forest}
\caption{Two benchmarking metrics in HD computing and some possible ways to improve these metrics.}\label{f:tree}
\end{figure*}
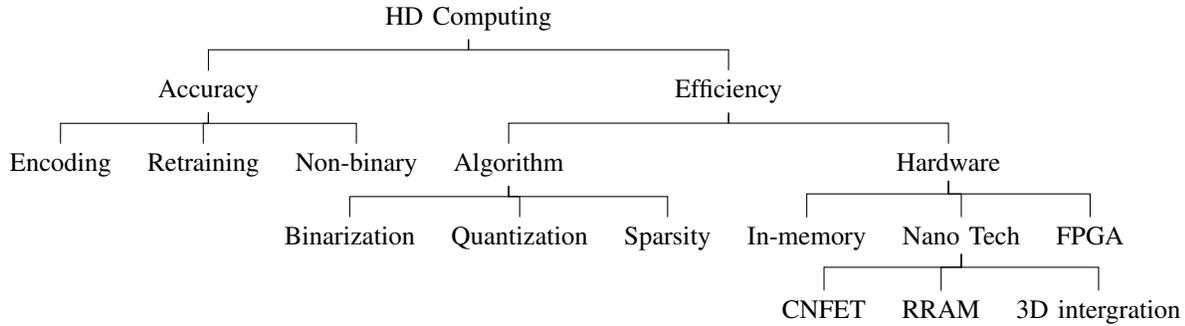

\subsubsection{$N$-gram-based Encoding} \textcolor{black}{The method of mapping $N$-gram statistics into hypervectors was proposed in \cite{joshi2016language}. First random level hypervectors are generated. Then the feature values are} obtained by permuting these level hypervectors in this encoding method. For example, the level hypervector $\mathbf{\bar{L}}_i$ corresponding to the $i$-th feature position is rotationally permuted by $(i-1)$ positions, where $1 \leq i \leq N$. We can get the final encoded hypervector $\mathbf{H}$ by Eq. (\ref{eq:ngram}). Such an encoding process is illustrated in Fig. \ref{f:ngram}.
\begin{equation}\label{eq:ngram}
 \begin{aligned}
& \mathbf{H} = \mathbf{\bar{L}}_1 \oplus \rho \mathbf{\bar{L}}_2 \oplus \cdots \oplus \rho ^{N-1} \mathbf{\bar{L}}_N,\\
& \mathbf{\bar{L}}_i  \in \{\mathbf{L}_1,\mathbf{L}_2,\cdots, \mathbf{L}_m \}, \; \text{where} \; 1 \leq i \leq N.
\end{aligned}
\end{equation}

\begin{remark}
As stated in \cite{imani2018hierarchical}, for speech recognition, the $N$-gram-based encoding method achieves lower accuracy than record-based counterpart. This encoding method is also used to address data types of letters, such as language recognition \cite{rahimi2016robust} and DNA sequencing \cite{imani2018hdna}. 
\end{remark}


\subsection{Benchmarking Metrics in HD Computing}
In HD computing, there is always a tradeoff between accuracy and efficiency, e.g., see \cite{kleyko2018classification}. As shown in Fig. \ref{f:tree}, a large amount of work has been \textcolor{black}{carried out} to improve the classification accuracy, energy efficiency, or both at the same time. 
\subsubsection{Accuracy}
In terms of accuracy, the encoding method plays a significant role since each encoding may not be efficient for different types of data. Good encoding for HD to achieve high accuracy is hard \cite{gayler2004vector}. In this sense, an appropriate choice of encoding method can improve the accuracy. Efficient encoding approaches have been presented in \cite{rahimi2018efficient}. The approach in \cite{imani2018hierarchical} integrates different encoding methods together to achieve higher accuracy at the expense of hardware area. Compared to single-pass training, retraining iteratively improves the training accuracy \cite{imani2017voicehd}. Thus the classification accuracy is improved by using a more accurately trained model. Moreover, using binary hypervectors may degrade the accuracy. Hence with enough resources, non-binary models can be used to achieve high accuracy.

\subsubsection{Efficiency}
For efficiency, improvements mainly focus on algorithm and hardware characteristics. From the algorithm perspective, dimension reduction is the most natural way to realize efficiency. Simulations show that slightly reducing the dimensionality of hypervectors, the classification accuracy still remains in an acceptable range but saves hardware resources \cite{imani2019quanthd}. Binarization, which refers to employing binary hypervectors instead of non-binary model, accelerates computation and reduces hardware resources \cite{imani2019binary}. The precision is degraded by quantizing the non-binary HD model. QuantHD has been proposed in \cite{imani2019quanthd} to achieve higher efficiency with minimal impact on accuracy. Sparsity was introduced in HD computing in the framework of BSDC \cite{rachkovskij2001representation}. Tradeoff between dense and sparse binary vectors has been presented in \cite{kleyko2018classification}. By introducing the concept of sparsity to hypervector representation, \cite{imani2019sparsehd} proposes a novel platform, SparseHD, which reduces inference computations and leads to high efficiency. From the hardware perspective, HD computing involves a large number of bit-wise operations, as well as the same computation flow for different HD applications, making FPGA a nice platform for hardware acceleration \cite{salamat2019f5}. Moreover, as proposed in \cite{in-mem2019}, combining HD computing with the concept of in-memory computing, which is featured as RAM storage and parallel distribution, may create opportunities for HD acceleration. Additionally, several emerging nanotechnologies, including carbon nanotube field-effect transistors (CNFETs) \cite{rahimi2018hyperdimensional}, resistive RAM (RRAM) \cite{rahimi2017high}, and monolithic 3D integration \cite{wu2018hyperdimensional}, have demonstrated  implementations of HD computing at high speed \cite{rahimi2018hyperdimensional}. \textcolor{black}{Dimensionality reduction has been evaluated in an actual prototyped system using vertical RRAM (VRRAM) in-memory kernels in \cite{li2016hyperdimensional}.}

\section{Applications in HD Classification}\label{s:4}
In what follows, some classical HD computing applications \textcolor{black}{in classification tasks} as well as several novel design approaches that can balance tradeoff of accuracy and efficiency are described. \textcolor{black}{They are categorized based on their input data types, namely letters, signals and images.}

\subsection{\textcolor{black}{Letters}}
\subsubsection{European Language Recognition Using HD Computing}
\textcolor{black}{HD computing for European language recognition was first explored by \cite{joshi2016language}.} Literature \cite{rahimi2018hyperdimensional} presents an HD computing nanosystem, which implements HD operations based on emerging nanotechnologies---CNFETs, RRAM and 3D integration---offering large arrays of memory and resulting in \textcolor{black}{reduction of energy consumption.} From its three-letter sequence called \textit{trigrams}, such a nanosystem can identify the language of a given sentence \cite{rahimi2018hyperdimensional}. Define a profile by a histogram of trigram frequencies in the unclassified text. The basic idea is to compare the trigram profile of a test sentence with the trigram profiles of 21 languages, and then find the target language which has the most similar trigram profile \cite{joshi2016language}.

\begin{figure*} 
\centerline{\includegraphics[width=\linewidth]{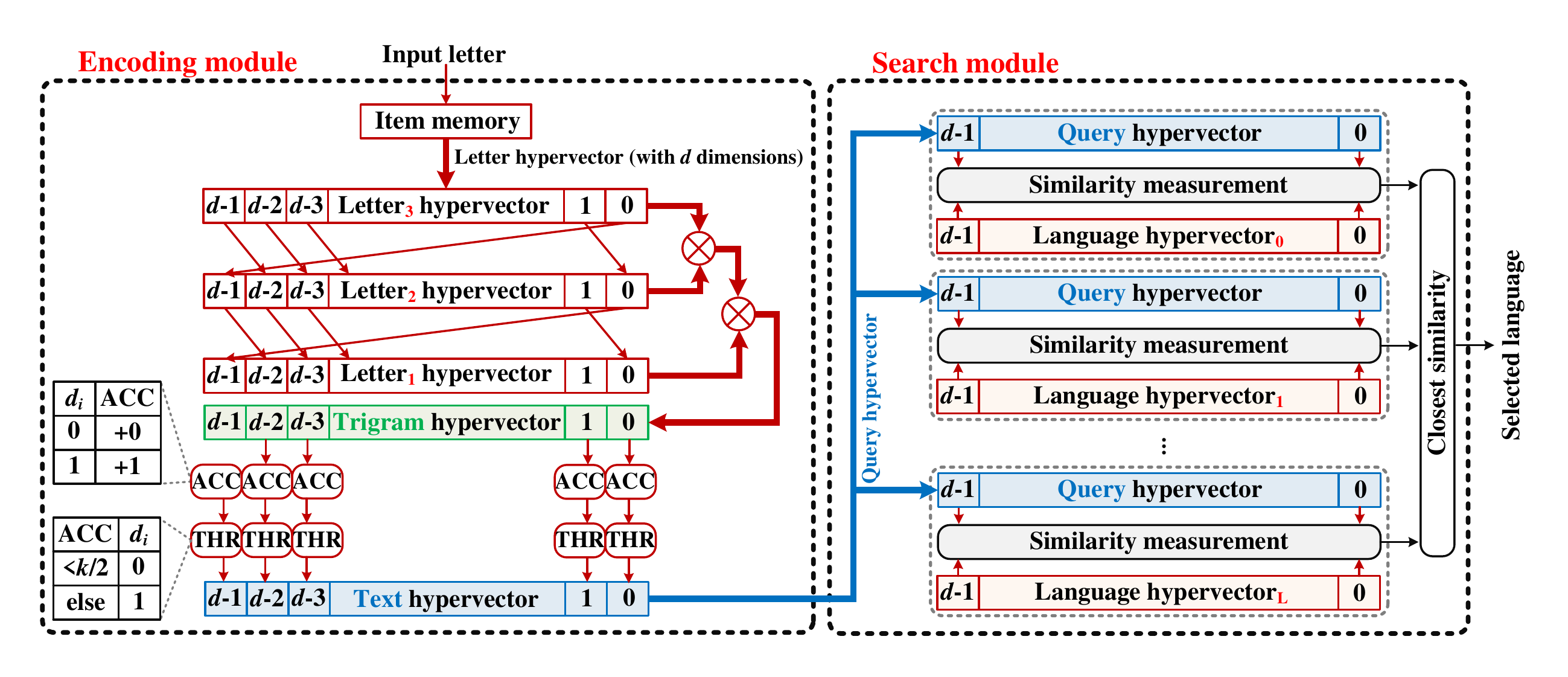}}
\caption{The architecture for \textcolor{black}{language recognition} with HD computing \cite{rahimi2016robust,rahimi2018hyperdimensional}.}
\label{f8}
\end{figure*}

\begin{itemize}
    \item \textbf{Baseline.} Scan through the text and count the trigram to compute a profile. A total of $27^3=19,683$ trigrams are possible for the 26 letters and the space. Thus the trigram counts can be encoded into a 19,683-dimensional vector and such vectors can be compared to find the language with the most similar profile. However, this straightforward and simple approach generalizes poorly. Specifically, compared to trigrams, higher-order $N$-grams will have higher complexity. For example, the number of possible pentagrams is $27^5 = 14,348,907$. 
    \item \textbf{HD classification algorithm.} \textcolor{blue}{\textbf{\textit{1).}}} Choose a set of 27 letter hypervectors randomly, serving as the seed hypervector. Note that all training and test data employ the same seeds. In this design, the dimensionality is selected to be 10,000. \textcolor{blue}{\textbf{\textit{2).}}} Generate trigram hypervectors with permutation and multiplication. For example, let $(a,b,c)$ represent a trigram. Then rotate the hypervector $A$ twice, hypervector $B$ once, and use hypervector $C$ with no change, and then multiply them component by component as described in Eq. (\ref{eq8}). \textcolor{blue}{\textbf{\textit{3).}}} The target profile hypervector is then the sum of all the trigram hypervectors in the text. \textcolor{blue}{\textbf{\textit{4).}}} Compare the profile of a test sentence to the language profiles, and return the most similar one as the classification result. 
\end{itemize}
 
Compared to the baseline algorithm, the HD algorithm generalizes better to any $N$-gram size when 10,000-dimensional hypervectors are used. 



The HD classification hardware architecture for language recognition using trigrams proposed in \cite{rahimi2016robust} is shown in Fig. \ref{f8}. Two main modules are implemented. They include the encoding module and the search module. \textcolor{blue}{\textbf{\textit{1).}}} The encoding module takes a stream of letters as the input. Each letter is mapped to the HD space and its corresponding randomly generated hypervector is stored in the item memory. Here it addresses the trigrams where each group of three hypervectors produces a trigram hypervector. Accumulate those trigram hypervectors and perform the majority operation using the threshold to generate a text hypervector. \textcolor{blue}{\textbf{\textit{2).}}} During the training phase, a total of 21 text hypervectors are trained as the learned class hypervectors and are stored in the \textcolor{black}{associative memory} in the search module. During the testing phase, the encoding module generates the text hypervector as a query hypervector. This query hypervector is then broadcast to the search module and compared to the stored class hypervectors to predict the language label, which has the closest similarity. As listed in Table \ref{t1}, the HD classifier achieves $96.70 \%$ accuracy.

Using the same architecture shown in Fig. \ref{f8}, and combining with the emerging nanotechnologies---CNFETs, RRAM and their monolithic 3D integration---the HD computing hardware implementation achieves classification accuracy up to $98 \%$  for over $>20,000$ sentences \cite{rahimi2018hyperdimensional}. 


\begin{figure*}[ht] 
\centerline{\includegraphics[width=0.85\linewidth]{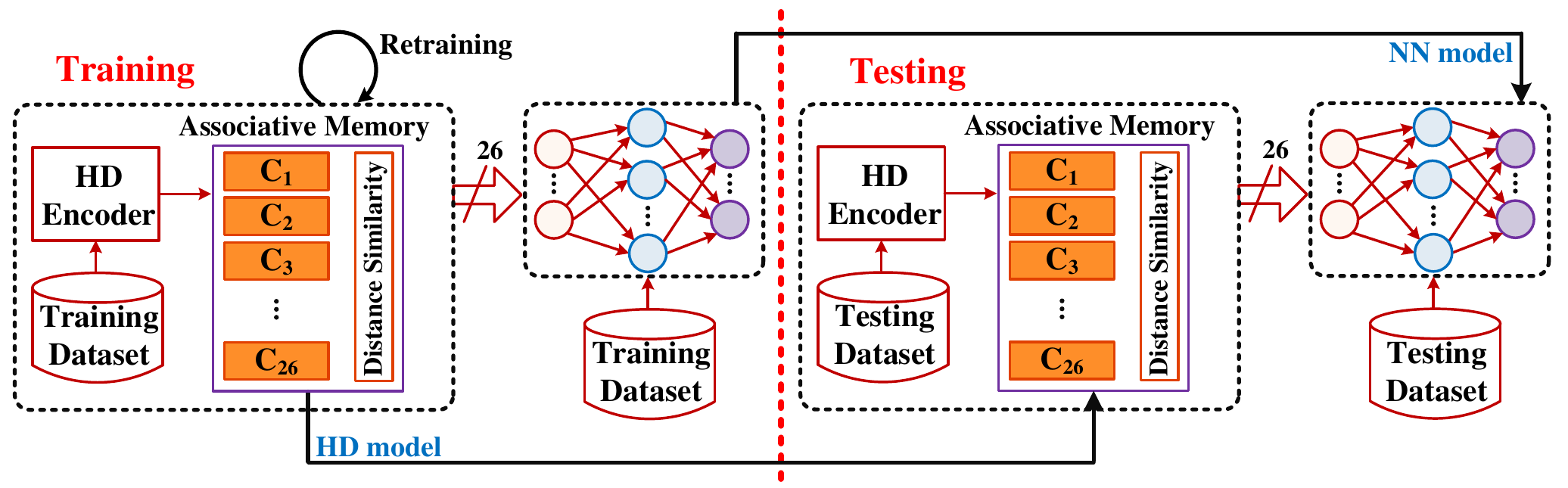}}
\caption{VoiceHD+NN flow for training and testing \cite{imani2017voicehd}.}\label{f11}
\end{figure*}

\begin{figure*}[htbp] 
\centerline{\includegraphics[width=0.98\linewidth]{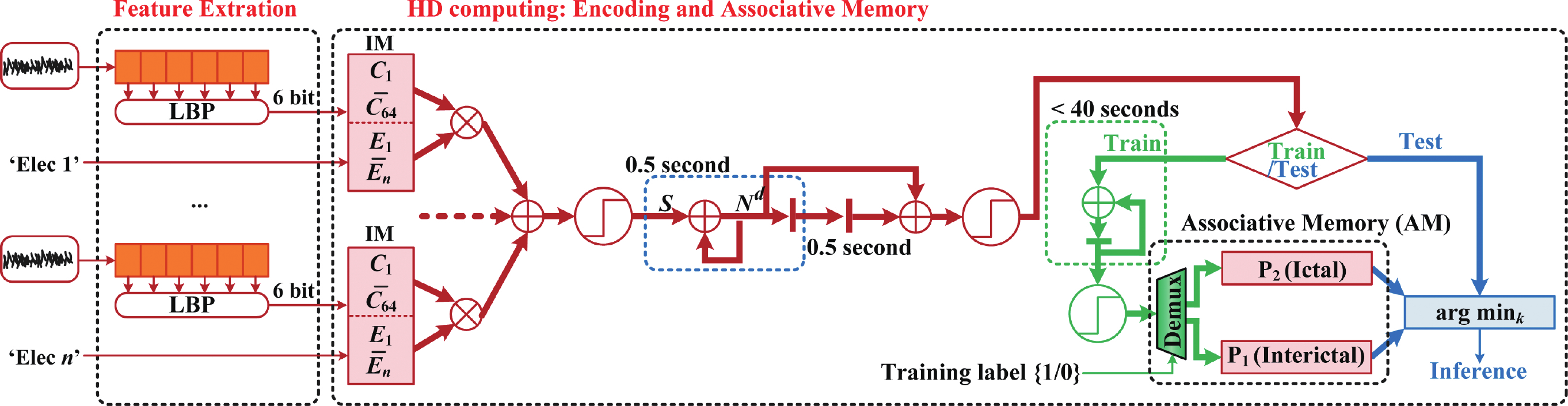}}
\caption{The architecture for Laelaps with HD computing to detect and alarm seizure \cite{burrello2019laelaps}.}
\label{f2}
\end{figure*}

\subsection{Signals}

\subsubsection{HD Classification for Speech Recognition}
The development of the Internet of Things (IoT) has motivated the market need for speech recognition. Though deep neural networks (DNNs) have been widely used for speech recognition, it requires expensive hardware and high energy consumption. This has inspired research for speech recognition based on HD computing which can achieve fast computation and energy efficiency. 

In \cite{imani2017voicehd}, VoiceHD, a new speech recognition technique, is proposed for classifying 26 letters from the spoken dataset. At the beginning, the voice signal is transformed to the frequency domain, which contains $N$ frequency ID channels and $M$ levels. Then VoiceHD maps these ID and level information into random hypervectors stored in the item memory. Combining these hypervectors, in the training phase, VoiceHD encoding module generates the learned patterns corresponding to 26 hypervectors that are stored in the associative memory. In the testing phase, VoiceHD uses the same encoding module to generate the query hypervector, which is broadcast to the associative memory. Comparing the query hypervector with the stored 26 class hypervectors, the hypervector with maximum similarity is retrieved to predict the letter. Here, dimensionality $d$ of the hypervectors is $10,000$. 
 

Researchers tested their VoiceHD design over Isolet dataset \cite{isolet}, where a total of 150 subjects spoke the name of each letter of the alphabet twice. The key findings are as follows: \textcolor{blue}{\textbf{\textit{1).}}} Varying the value of $M$, the number of levels of the amplitude between $-1$ and $1$, with $N$, the number of frequency bins, fixed at $617$, the recognition accuracy increases with increase in $M$. Note the encoding efficiency degrades with large $M>10$. The maximum accuracy reaches $88.4 \%$ using $M=10$. \textcolor{blue}{\textbf{\textit{2).}}} To improve the classification accuracy, researchers retrain the associative memory by modifying the trained class hypervectors. The accuracy can be improved to $93.8 \%$. \textcolor{blue}{\textbf{\textit{3).}}} Combining VoiceHD with a small neural network, the corresponding VoiceHD + NN flow is shown in Fig. \ref{f11}. Such a small NN has three layers. There are 26 neurons in the first layer, 50 neurons in the hidden layer and another 26 neurons in the last layer. The classification accuracy can be improved to be $95.3 \%$. \textcolor{blue}{\textbf{\textit{4).}}} Compared to the pure NN with $93.6 \%$ classification accuracy, VoiceHD and VoiceHD+NN show $4.6 \times$ and $2.9\times$ faster training speed, $5.3 \times$ and $4.0 \times$ faster testing speed, and $11.9 \times$ and $8.6 \times$ higher energy efficiency, respectively.



\subsubsection{Seizure Detection Using HD Computing}

The Laelap algorithm, which utilizes local binary pattern (LBP) codes to conduct the feature extraction  from iEEG signals, has been proposed in \cite{burrello2019laelaps} for seizure prediction. Here HD computing is applied to capture the statistics of the time-varying LBP codes for all the electrodes. Fig. \ref{f2} illustrates the complete processing chain. \textcolor{blue}{\textbf{\textit{1).}}} Since the down-sampling frequency is 512 Hz, thus every one second (1s) data contains 512 samples. 
Among these samples, the sampled iEEG signals are encoded to 6-bit LBP codes. This completes the feature extraction part. \textcolor{blue}{\textbf{\textit{2).}}} It utilizes record-based encoding, \textcolor{black}{ where two types of hypervectors are randomly generated.} Specifically, each LBP code is transformed to a $d$-dimensional hypervector $C_i$, while the hypervectors $E_i$ are used to represent the corresponding electrode name. For every new sample, the hypervectors $E_i$ and $C_i$ are bound together to form a \textcolor{black}{composite} hypervector $S=[C_1 \oplus E_1 +\cdots C_{n} \oplus E_{n}]$, where $n$ is the number of electrodes for a specific patient. Then the histogram of LBP codes $H$ is computed for a moving window of 1s with 0.5s overlap. Therefore the composite hypervector $H=[S^{1}+S^{2}+\cdots+S^{512}]$ is updated every 0.5s. \textcolor{blue}{\textbf{\textit{3).}}} For learning, two prototype hypervectors $P_1$ and $P_2$ should be trained. For interictal prototype vector $P_1$, all $H$ computed over 30s should be accumulated and normalized to be stored in the associative memory. Depending on the seizure's duration, the ictal prototype vector $P_2$ is generated using all $H$ over an ictal state, which may last 10s to 30s. \textcolor{blue}{\textbf{\textit{4).}}} For classification, comparing $P_k$ with a query $H$, the label is updated every 0.5s with the shortest Hamming distance $\text{Ham}(H, P_k)$, where $k=1,2$. \textcolor{blue}{\textbf{\textit{5).}}} The algorithm also generates the seizure alarm. In postprocessing, if the last 10 labels all indicate $P_2$ ($t_c=10$) and the distance score $\Delta > t_r$, then the seizure alarm is generated. 

The evaluation shows the Laelaps algorithm outperforms other machine learning methods, such as SVM, in terms of energy efficiency. It is worth noting that many simpler seizure detection and prediction algorithms have been proposed in the literature \cite{zhang2014seizure,zhang2015low,zhang2015seizure,zhang2015seizure2,parhi2019discriminative}. A fair comparison of classifier accuracy between HD and traditional classification needs to be explored in future.

\subsubsection{Quantization in HD Computing}
\textcolor{black}{In dealing with signals,} HD computing usually makes use of floating point models to improve the classification accuracy at the cost of high computation cost. In \cite{imani2019quanthd}, QuantHD is proposed as a quantization of HD model, which projects the trained non-binary hypervectors to a binary or ternary model, with elements in  $\{0,1 \}$ or $\{-1,0,+1\}$, to represent class hypervectors. To compensate the accuracy degradation caused by quantization, a retraining approach is used where an iteration number of $30$ is pre-defined. The similarity check is no longer cosine metric (non-binary model), but Hamming distance (binary model) or dot product (ternary model). Compared to the existing binarized HD computing, such QuantHD improves on average $17.2\%$ accuracy with a similar computation cost.

\subsubsection{HD Computing Using Model Compression}
As a mathematical framework, HD computing can be an alternative for machine learning problems. This was envisioned in \cite{gallant2013representing}. Due to the high dimensionality, the inference of HD computing is quite expensive, especially when it is applied to the embedded devices with limited resources. For example, the memory is limited. Therefore, reducing the high dimensionality of hypervectors without sacrificing the accuracy has been investigated in \cite{morris2019comphd}. Thus, CompHD is a general method that compresses the model size with the minimal loss of accuracy. The addressed hypervectors are in $\{-1,1 \}^d$. Instead of Hamming distance, the similarity metric in CompHD is cosine similarity.
\begin{figure}[ht] 
\centering
\includegraphics[width=\linewidth]{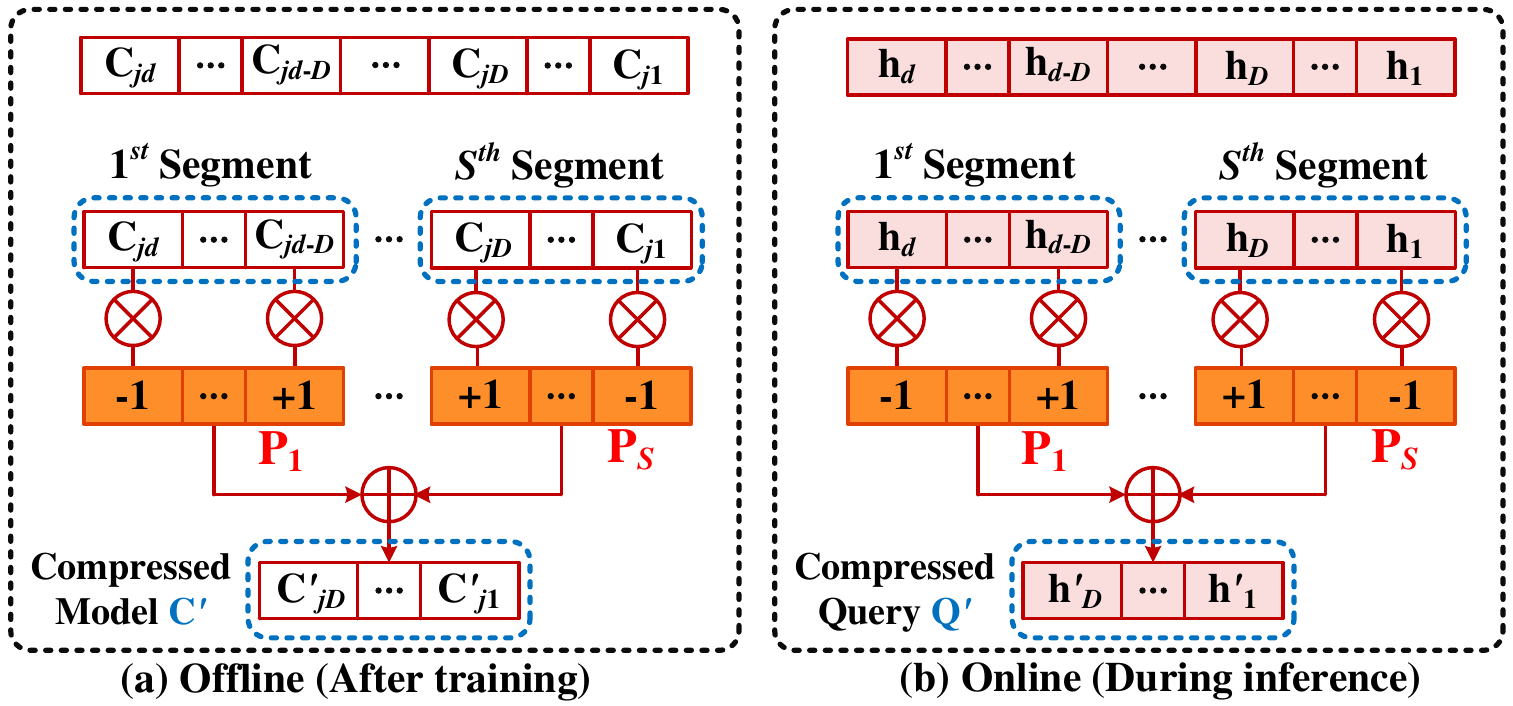}
\caption{CompHD for (a) an HD model and (b) a query data  \cite{morris2019comphd}.}\label{f12}
\end{figure}

To reduce the HD model size, it is natural to use low dimensional hypervectors. However, experimental results of three practical applications using different dimensionalities in HD classification show that the efficiency is improved by reducing model size at the cost of accuracy. 

To maintain high accuracy when reducing the dimensionality, the proposed CompHD employs the architecture shown in Fig. \ref{f12}. With no reduction in model size, $C_i$ represents the class hypervector, $Q$ represents the query hypervector, where $ 1 \leq i \leq k$. In CompHD, class hypervectors and query hypervectors are compressed, which means the original hypervectors are divided into $s$ segments. To store most of the information in original hypervectors with the full size, using Hadamard method \cite{hadamard}, CompHD generates $P_1, P_2, \cdots, P_s$, which are in $\{-1,1\}^D$ and are orthogonal to each other, where $D=d/s$. Specifically, the compressed class hypervector $C'$ and query hypervector $Q'$ are calculated using multiplication and addition in HD as described by Eq. (\ref{eq10}). By doing so, only little information is lost when we compress the model size, and high accuracy can be maintained.
\begin{equation}\label{eq10}
 \begin{aligned}
 C'=\sum_{i=1}^{s}P_iC^i, \qquad Q'=\sum_{i=1}^{s}P_iQ^i\\
\end{aligned}
\end{equation}
Their evaluation shows that, compared to the original HD classification that purely reduces the dimensionality with the compression factor $s=20$, the classification accuracy for the three applications is still in an acceptable range. In particular, maintaining the same accuracy as the original, CompHD can on average reduce model size by $69.7 \%$ while still achieving $74\%$ energy improvement and $4.1 \times$ execution time speedup in the context of activity recognition, gesture recognition and valve monitoring applications \cite{morris2019comphd}. Therefore, CompHD is suitable for low-power IoT devices to achieve higher efficiency with a comparable accuracy.

\begin{figure*}[ht] 
\centering
\includegraphics[width= \linewidth]{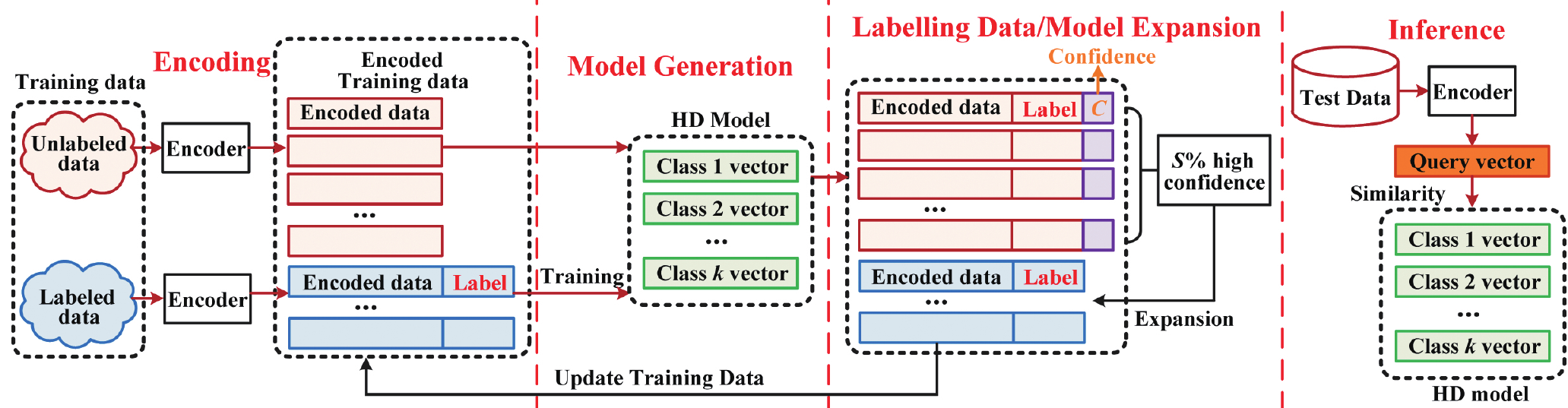}
\caption{Overview of SemiHD framework supporting self-training in HD space \cite{imani2019semihd}.}\label{f16}
\end{figure*}

\subsubsection{Adaptive Efficient Training for HD Computing}
Single-pass training leads to low accuracy. To improve this, iterative training might be one efficient solution. However, a lack of controllability of training iterations in HD classification may result in slow training or divergence. To solve this training issue, \cite{imani2019adapthd} proposes a retraining approach, AdaptHD.

The basic idea is illustrated as follows: \textcolor{blue}{\textbf{\textit{1).}}} Conduct the initial training by using binary hypervectors to generate the non-binary class hypervectors. \textcolor{blue}{\textbf{\textit{2).}}} Retrain the class hypervectors by looking at the similarity of each trained class hypervectors ($C$) with the training hypervector ($H$).  Update the model using Eq. (\ref{eq11}) if the current training hypervector leads to a mislcassification error. Otherwise there is no change. For example, there is a mismatch if $H_i$ is supposed to belong to $C_\text{correct}$ but is classified as $C_\text{wrong}$, where $C_\text{correct}$ and $C_\text{wrong}$ denote different class hypervectors and $H_i$ represents the $i$th training hypervector. \textcolor{blue}{\textbf{\textit{3).}}} After convergence, which means the last three iterations of retraining show less than $0.1\%$ accuracy change, then binarize the final trained model for inference. 
\begin{equation}\label{eq11}
\left\{
 \begin{aligned}
 C_\text{wrong}=C_\text{wrong}-\alpha H_i, \\C_\text{correct}=C_\text{correct}+\alpha H_i.\\
\end{aligned}
\right.
\end{equation}

Insights are gained by their results: \textcolor{blue}{\textbf{\textit{1).}}} Small $\alpha$ needs more iterations to get the near best accuracy. The smooth curve indicates small $\alpha$ is better for fine-tuning. \textcolor{blue}{\textbf{\textit{2).}}} Large $\alpha$ gets to the near best accuracy much faster, but its high fluctuation may lead to divergence. Based on these two findings, AdaptHD uses large $\alpha$ first to get the near best accuracy faster, then changes to smaller $\alpha$ for fine-tuning until convergence. This is similar to adjusting the step size in the normalized least mean square (LMS) algorithm \cite{haykin2014adaptive}. AdaptHD offers three types of adaptive methods:
\begin{itemize}
    \item Iteration-dependent AdaptHD. The change of value $\alpha$ depends on iterations. In the beginning, $\alpha$ starts with a large $\alpha_{max}$. The learning rate $\alpha$  changes based on the average error rate in the previous $\beta$ iterations. If error rate decreases, indicating convergence, then use smaller $\alpha$; otherwise, increase $\alpha$.
    \item Data-dependent AdaptHD. The value $\alpha$ differs in a certain iteration for all data points, and it changes depending on the similarity of the data point with the class hypervectors. Large distance uses large $\alpha$ to reduce the difference.
    \item Hybrid AdaptHD. Combining the two models, hybrid AdaptHD can achieve high accuracy as iteration-dependent AdaptHD and fast speedup as data-dependent AdaptHD. 
\end{itemize}

The evaluation shows that, compared to the existing HD algorithm, their hybrid AdaptHD can achieve $6.9 \times $ speedup and $6.3 \times $ energy-efficiency improvement.

\subsubsection{A Binary Framework for HD Computing}
Generally speaking, HD classification using binary hypervectors shows lower accuracy but higher energy efficiency than non-binary ones. This is because the non-binary framework makes use of the costly cosine similarity rather than the hardware-friendly Hamming distance metric. In \cite{imani2019binary}, BinHD uses three main blocks, encoding, associative search and counter modules, dealing with binary hypervectors. Their evaluation shows that, over four practical applications, the proposed BinHD can reach $12.4 \times$ and $6.3 \times $ energy efficiency and speedup in training process, while $13.8 \times$ and $9.9 \times $ during inference, compared to the state-of-art HD computing algorithm with a comparable classification accuracy.

\begin{figure*}[ht] 
\centering
\includegraphics[width= \linewidth]{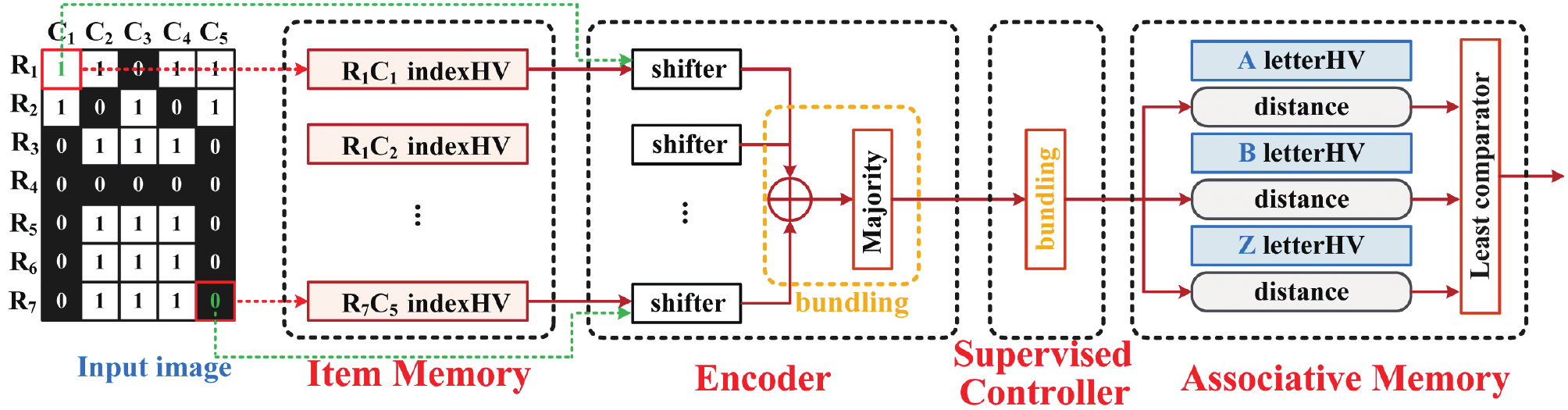}
\caption{Block diagram of the HD Character Recognition System  \cite{manabat2019performance}.}\label{f17}
\end{figure*}

\subsubsection{HD Computing for Semi-Supervised Learning}
In \cite{imani2019semihd}, SemiHD  has been proposed as a self-training or self-learning approach for semi-supervised learning, where the training data is composed of a small portion of labeled data and a large portion of unlabeled data.

The SemiHD framework is depicted in Fig. \ref{f16} and the flow is illustrated as follows. \textcolor{blue}{\textbf{\textit{1).}}} Encode all the data points, labeled and unlabeled, into HD space with $d=10,000$ dimensions. 
\textcolor{blue}{\textbf{\textit{2).}}} Start training from the labeled data to generate $k$ hypervectors, each representing one class. 
\textcolor{blue}{\textbf{\textit{3).}}} Predict the label for unlabeled data points. Labeling is performed by checking the similarity of unlabeled data with all the class hypervectors, and return the label which shows the highest similarity. 
\textcolor{blue}{\textbf{\textit{4).}}} Select and add $S\%$ of unlabeled data with highest confidence to labeled data, where $S$ is defined as the expansion rate. In \cite{imani2019semihd}, typically $S=5$.
\textcolor{blue}{\textbf{\textit{5).}}} Redo the training task based on the expanded labeled data. Such iterative process stops when the accuracy does not change more than 0.1\%. 
\textcolor{blue}{\textbf{\textit{6).}}} Once the model has already been trained, perform the inference task by comparing the similarity of each test data with the trained model, to return the label with maximum similarity.

Their evaluation shows that the SemiHD can on average improve the classification of supervised HD by $10.2\%$. Additionally, compared to the best CPU implementation, the FPGA counterpart of SemiHD offers $7.11 \times$ faster speed and $12.6 \times$ energy efficiency.

\subsubsection{HD Computing for Unsupervised Learning}
HD computing has also been used in several unsupervised applications. See \cite{kanerva2000random,recchia2015encoding,kleyko2019distributed,bandaragoda2019trajectory,hersche2020evolvable}.
\subsection{\textcolor{black}{Images}}
\subsubsection{HD Classification for Character Recognition}
HD classification has been used for character recognition in \cite{kleyko2016holographic} and later in \cite{manabat2019performance}. As shown in Fig. \ref{f17}, the input image is composed of $7 \times 5=35$ pixels. Each pixel has two possible values, that is 0 or 1, representing black or white. 
\textcolor{blue}{\textbf{\textit{1).}}} Encode each pixel to a binary hypervector (indexHV). Totally 35 orthogonal indexHVs are stored in the item memory.
\textcolor{blue}{\textbf{\textit{2).}}} Based on HoloGN encoding\textcolor{black}{---an encoding method proposed in \cite{kleyko2016holographic} to address image data using HD computing---}the indexHV is shifted depending on the pixel value. Accumulate all 35 indexHVs and perform a majority rule by a thresholding block to generate a holoHV for one input image. \textcolor{blue}{\textbf{\textit{3).}}} The supervised controller will only be activated when this HD system conducts supervised learning. Otherwise, the system conducts the one-shot learning. The supervised controller \textcolor{black}{accumulates} the holoHVs for the same class and employs the thresholding block and generates the letterHV to be stored in the associative memory. The total number of letterHVs is 26. \textcolor{blue}{\textbf{\textit{4).}}} During the test phase, the query hypervector is generated following the same module with test data. Then the similarity of each query hypervector is computed for all trained letterHVs to find the most similar class. 

Results in \cite{manabat2019performance} show that HD computing \textcolor{black}{performs well for} character recognition. Further optimization for HD computing may be conducted by reducing the dimensionality and increasing the input image size. The results also show that HD computing offers great robustness against noise. The system of 4,000-bit hypervectors achieves comparable average accuracy to its 12,000-bit counterpart at $0\%$ distortion, and achieves an average accuracy of $89.94\%$ with $14.29 \%$ distortion.

\begin{table*}
\centering
\caption{Summary of the strategies used in HD computing for accuracy and efficiency improvement.}
\setlength{\leftskip}{-69pt}
\vspace{4pt}
\begin{threeparttable}
\begin{tabular}{c||c|c|c|c|c|c|c|c|c}
\hline
\multirow{2}{*}{ \textbf{Applications} } & \multirow{2}{*}{ \textbf{Encode}\tnote{*} } &  \multicolumn{3}{c|}{ \textbf{Model Type}\tnote{**}} & \multirow{2}{*}{ \textbf{Platform}\tnote{$\Delta$} } & \multirow{2}{*}{ \textbf{Accuracy}\tnote{$\heartsuit$}} & \multirow{2}{*}{ \textbf{Acceleration}\tnote{$\clubsuit$}} & \multirow{2}{*}{ \textbf{Motivation} } & \multirow{2}{*}{ \textbf{Application} } \\
\cline{3-5}
& & \textbf{base/level} & \textbf{train} & \textbf{test} & & & & & \\
\hline
\hline
QuantHD \cite{imani2019quanthd} & 1 & B/B & B/T & B/T & F, C, G & Re, shuffle & Q, DR, F & speedup+accuracy &  speech, activity, face, phone position\\
VoiceHD \cite{imani2017voicehd} & 1 & B/B & B & B & C &  NN, Re & DR, B & Replace deep learning & speech \\
CompHD \cite{morris2019comphd} & 3 & P & N & N & F & N & DR, Comp & DR without accuracy loss & activity, gesture, valve monitoring \\
AdaptHD \cite{imani2019adapthd} & 1 & B/B & N & B & C & Re, N & B, Adapt & accuracy+short time Re & speech, face, activity, Cardiotocograms \\
BinHD \cite{imani2019binary} & 1 & B/B & B & B & C & Re & B & speedup & speech, face, activity, Cardiotocograms \\
SemiHD \cite{imani2019semihd} & 1 & B/B & B & B & F, C & Re, N &  DR, B & Replace \textcolor{black}{deep learning}  & 17 popular datasets \cite{triguero2015self}\\
Language \cite{rahimi2016robust} & 2 & B & B & B & \textcolor{black}{F} & \backslashbox{ }{ } & B & \textcolor{black}{energy saving + robustness} & language recognition\\
 Character \cite{manabat2019performance}  & 3 & B & B & B & \backslashbox{ }{ } &  Binary & DR &  data classification in IoT & character recognition\\
Laelap \cite{burrello2019laelaps} & 1 & B & B & B & C, G & \backslashbox{ }{ } & \backslashbox{ }{ } & energy efficiency & seizure detection \\
\hline
\end{tabular}
 \begin{tablenotes}
        \footnotesize
        \item[*] three encoding methods. 1: record-based encoding, 2: $N$-gram-based encoding, 3: a novel method.  
        \item[**] symbol ``/'' is used in record-based encoding. B: binary, P: bipolar, T: Ternary, N: non-binary.
        \item[$\Delta$] implementation platforms. F: FPGA, C: CPU, G: GPU. 
        \item[$\heartsuit$] strategies for accuracy improvement. Re: retraining, N: non-binary model, NN: neural network.
        \item[$\clubsuit$] strategies for efficiency improvement. DR: dimension reduction, Q: quantization, B: binarization, F: FPGA, Comp: compression, Adapt: adaptive.
      \end{tablenotes}
    \end{threeparttable}\label{t:s}
\end{table*}

\begin{table*}[htbp]
  \centering
    \begin{threeparttable}
    \caption{Partial List of applications based on HD computing\tnote{\textcolor{blue}{$\clubsuit$}} in \cite{rahimi2018hyperdimensional}.}
    \begin{tabular}{c||c|c|c|c}
   \hline
    Applications  &  {\textcolor{black}{Inputs (\#)}\tnote{*} } & {Classes (\#)\tnote{**} } &  {HD (\%) } &  {Baseline (\%)} \\
     \hline
      \hline
    Language recognition \cite{rahimi2016robust,joshi2016language}  & 1     & 21    & 96.70\% & 97.90\% \\
   \hline
    Text categorization \cite{najafabadi2016hyperdimensional}    & 1     & 8     & 94.20\% & 86.40\% \\
    \hline
    Speech recognition \cite{imani2017voicehd}   & 1     & 26    & 95.30\% & 93.60\% \\
     \hline
    EMG gesture recognition \cite{rahimi2016hyperdimensional}  & 4     & 5     & 97.80\% & 89.70\% \\
      \hline
    Flexible EMG gesture recognition \cite{moin2018emg} & 64    & 5     & 96.60\% & 88.90\% \\
     \hline
    EEG brain-machine interface \cite{rahimi2017hyperdimensional}  & 64    & 2     & 74.50\% & 69.50\% \\
    \hline
    ECoG seizure detection \cite{burrello2018one}  & 100   & 2     & 95.40\% & 94.30\% \\
    \hline
    \textcolor{black}{DNA sequencing} \cite{imani2018hdna}  & 1   &   \backslashbox{ }{ }    & 99.74\% & 94.53\% \\
    \hline
    \textcolor{black}{Character  recognition \cite{manabat2019performance}} & 1 & 10 & 89.94\% &  \backslashbox{ }{ } \\
    \hline
    \end{tabular}\label{t1}
    \begin{tablenotes}
        \footnotesize
        \item[*]  \textcolor{black}{represents the number of input data.}
        \item[**] \textcolor{black}{represents the number of class hypervectors to be trained and stored in the associative memory.}
        \item[\textcolor{blue}{$\clubsuit$}] \textcolor{black}{Other works, like \cite{rasanen2014modeling,rasanen2015sequence,kleyko2015imitation,yilmaz2015connectionist,kleyko2017modality,montone2017hyper,kleyko2018hyperdimensional,mitrokhin2019learning}, are not listed in this table.}
      \end{tablenotes}
      \end{threeparttable}
\end{table*}%

\subsection{Summary}

As mentioned above, HD computing shows great potential in dealing with data in the form of signals \cite{burrello2019laelaps,rasanen2015generating,imani2017voicehd, burrello2018one,rahimi2017hyperdimensional}, letters \cite{najafabadi2016hyperdimensional,joshi2016language}, and images \cite{kleyko2016holographic,manabat2019performance,rahimi2017high}, as long as these can be transformed into the HD space. Such pre-processing may include feature extraction and encoding. Evaluation shows that HD computing achieves good results for seizure detection \cite{burrello2019laelaps, burrello2018one}. In addition, HD computing can also be combined with quantization technique to binarize HD model with minimal accuracy loss \cite{bosch2019qubithd}. Table \ref{t:s} offers more details about improvement strategies adopted in HD computing for accuracy and efficiency. As can been seen from Table \ref{t1}, HD computing offers an acceptable accuracy, but with quite high efficiency. \textcolor{black}{In some applications like DNA sequencing \cite{imani2018hdna}, HD computing outperforms other machine learning methods.}

\textcolor{black}{There still exist some interesting papers not discussed in detail in this review paper. Interested readers can refer to the following references, which include but are not limited to: \textcolor{blue}{\textit{1).}} Considering the security issue when IoT devices release the offload computation to the cloud, \cite{imani2019framework} illustrates how the proposed SecureHD accelerates efficiency with high security. \textcolor{blue}{\textit{2).}} To balance the tradeoff between efficiency and accuracy, QubitHD \cite{bosch2019qubithd} is proposed as a stochastic binarization algorthim to achieve comparable accuracy to the non-binarized counterparts. SparseHD \cite{imani2019sparsehd} takes advantage of the sparsity of the trained HD model for acceleration. }

HD computing is still in its infancy. Future directions may include but is not limited to:
\begin{itemize}
    \item \textcolor{black}{More cognitive tasks: Inspired by \cite{kleyko2018classification}, apart from the engineering aspect of HD computing, which is to solve classification tasks, more ``cognition'' aspects of HD computing should be explored. Such tasks include but are not limited to analogical reasoning, semantic generalization and relational representation.}
    \item Feature exaction and encoding method: Since HD computing cannot directly address data like signals and images, feature exaction is vital to representation of information. \textcolor{black}{For example, \cite{rasanen2015generating} partially deals with this by addressing the problem of mapping data to a high-dimensional space.}
    \item Similarity measurement: Though cosine similarity and Hamming distance are currently widely used, new metrics should be developed that are hardware-friendly and can lead to high accuracy.
    \item Multiple class hypervectors: Traditional classifiers use multi-dimensional features to train a classifier. Often ranking can be used to select a small number of features out of many features \cite{zhang2018muse}. It is possible that multiple class hypervectors, similar to multiple features in traditional classification, can be generated to represent a class in HD classification. Subsequently, multiple query hypervectors will need to be compared with their corresponding class hypervectors for each class. This is a topic for further research.
    \item \textcolor{black}{Accuracy improvement: Strategies like retraining should be explored to further improve the accuracy of HD computing.} 
    \item Hardware acceleration: Rebuilding the specific implementation for HD computing to store and manipulate a large amount of hypervectors may result in high speed and energy efficiency. \textcolor{black}{Moreover, inspired by \cite{kleyko2018classification}, which discusses tradeoffs related to the density of hypervectors, a choice between dense and sparse approaches should be accordingly made based on the application scenarios. For example, adopting sparse representation requires lower memory footprints.}
    \item General HD computing processor: Inspired by \cite{datta2019programmable}, addressing different types of data with only one general processor containing a large word-length ALU is of great interest.
    \item \textcolor{black}{Hybrid systems: Hybrid systems are partially based on HD computing and partially on conventional machine learning. Only a few examples exist so far \cite{alonso2020hyperembed,kleyko2019density,kleyko2017integer,anderson2017high}. Further research on this topic can be explored in future.}
\end{itemize}


\section{Conclusion}\label{s:5}
This paper has summarized the fundamental arithmetic operations for the emerging computing model of HD computing that might achieve high robustness, fast learning ability, hardware-friendly implementation, and energy efficiency. Mathematically, HD computing can be viewed as an alternative in dealing with machine learning problems. Though in its infancy, HD computing shows its potential to be used as a light-weight classifier for applications with limited resources. This model can achieve outstanding classification performance for certain problems like DNA sequencing. Balancing the tradeoff between accuracy and efficiency is an important area of research. Improvements include but are not limited to encoding, retraining, non-binary model and hardware acceleration. HD computing sometimes leads to outstanding classification accuracy, while sometimes achieves acceptable accuracy but high efficiency. Thus, users need to evaluate whether HD computing is suitable for their application. Additionally, HD computing can be used in applications such as seizure detection, speech recognition, character recognition and language detection. \textcolor{black}{More ``cognition" aspects of HD computing, including analogical reasoning, relationship representation and analysis, will need to be further developed in the future.}  

\section*{Acknowledgment}
This paper has been supported in parts by the NSF under grant number
CCF-1814759 and by the Chinese Scholarship Council (CSC). The authors thank all four reviewers for their numerous constructive comments and suggestions.

\ifCLASSOPTIONcaptionsoff
  \newpage
\fi

\footnotesize
\bibliographystyle{IEEEtran}
\bibliography{IEEEabrv,mybib}

\begin{thebibliography}{10}
\providecommand{\url}[1]{#1}
\csname url@samestyle\endcsname
\providecommand{\newblock}{\relax}
\providecommand{\bibinfo}[2]{#2}
\providecommand{\BIBentrySTDinterwordspacing}{\spaceskip=0pt\relax}
\providecommand{\BIBentryALTinterwordstretchfactor}{4}
\providecommand{\BIBentryALTinterwordspacing}{\spaceskip=\fontdimen2\font plus
\BIBentryALTinterwordstretchfactor\fontdimen3\font minus
  \fontdimen4\font\relax}
\providecommand{\BIBforeignlanguage}[2]{{%
\expandafter\ifx\csname l@#1\endcsname\relax
\typeout{** WARNING: IEEEtran.bst: No hyphenation pattern has been}%
\typeout{** loaded for the language `#1'. Using the pattern for}%
\typeout{** the default language instead.}%
\else
\language=\csname l@#1\endcsname
\fi
#2}}
\providecommand{\BIBdecl}{\relax}
\BIBdecl

\bibitem{kanerva1988sparse}
P.~Kanerva, \emph{{Sparse Distributed Memory}}.\hskip 1em plus 0.5em minus
  0.4em\relax MIT press, 1988.

\bibitem{smolensky1990tensor}
P.~Smolensky, ``Tensor product variable binding and the representation of
  symbolic structures in connectionist systems,'' \emph{Artificial
  intelligence}, vol.~46, no. 1-2, pp. 159--216, 1990.

\bibitem{plate1995holographic}
T.~A. Plate, ``Holographic reduced representations,'' \emph{IEEE Transactions
  on Neural networks}, vol.~6, no.~3, pp. 623--641, 1995.

\bibitem{kanerva1997fully}
P.~Kanerva \emph{et~al.}, ``Fully distributed representation,'' \emph{PAT},
  vol.~1, no.~5, p. 10000, 1997.

\bibitem{rachkovskij2001binding}
D.~A. Rachkovskij and E.~M. Kussul, ``Binding and normalization of binary
  sparse distributed representations by context-dependent thinning,''
  \emph{Neural Computation}, vol.~13, no.~2, pp. 411--452, 2001.

\bibitem{gayler1998multiplicative}
R.~W. Gayler, ``Multiplicative binding, representation operators \& analogy
  (workshop poster),'' 1998.

\bibitem{schlegel2020comparison}
K.~Schlegel, P.~Neubert, and P.~Protzel, ``A comparison of vector symbolic
  architectures,'' \emph{arXiv preprint arXiv:2001.11797}, 2020.

\bibitem{hersche2018exploring}
M.~Hersche, J.~d.~R. Mill{\'a}n, L.~Benini, and A.~Rahimi, ``Exploring
  embedding methods in binary hyperdimensional computing: A case study for
  motor-imagery based brain-computer interfaces,'' \emph{arXiv preprint
  arXiv:1812.05705}, 2018.

\bibitem{rahimi2017high}
A.~Rahimi, S.~Datta, D.~Kleyko, E.~P. Frady, B.~Olshausen, P.~Kanerva, and
  J.~M. Rabaey, ``High-dimensional computing as a nanoscalable paradigm,''
  \emph{IEEE Transactions on Circuits and Systems I: Regular Papers}, vol.~64,
  no.~9, pp. 2508--2521, 2017.

\bibitem{bryant2015computer}
R.~E. Bryant and D.~R. O'Hallaron, ``Computer systems: A programmer's
  perspective,'' 2015.

\bibitem{patyk2011comparison}
A.~Patyk-{\L}o{\'n}ska, M.~Czachor, and D.~Aerts, ``A comparison of geometric
  analogues of holographic reduced representations, original holographic
  reduced representations and binary spatter codes,'' in \emph{2011 Federated
  Conference on Computer Science and Information Systems (FedCSIS)}.\hskip 1em
  plus 0.5em minus 0.4em\relax IEEE, 2011, pp. 221--228.

\bibitem{olver2018applied}
P.~J. Olver and C.~Shakiban, \emph{Applied Linear Algebra}.\hskip 1em plus
  0.5em minus 0.4em\relax Springer, 2018.

\bibitem{datta2019programmable}
S.~Datta, R.~A. Antonio, A.~R. Ison, and J.~M. Rabaey, ``A programmable
  hyper-dimensional processor architecture for human-centric {IoT},''
  \emph{IEEE Journal on Emerging and Selected Topics in Circuits and Systems},
  vol.~9, no.~3, pp. 439--452, 2019.

\bibitem{widdows2015reasoning}
D.~Widdows and T.~Cohen, ``Reasoning with vectors: A continuous model for fast
  robust inference,'' \emph{Logic Journal of the IGPL}, vol.~23, no.~2, pp.
  141--173, 2015.

\bibitem{schmuck2019hardware}
M.~Schmuck, L.~Benini, and A.~Rahimi, ``Hardware optimizations of dense binary
  hyperdimensional computing: Rematerialization of hypervectors, binarized
  bundling, and combinational associative memory,'' \emph{ACM Journal on
  Emerging Technologies in Computing Systems (JETC)}, vol.~15, no.~4, pp.
  1--25, 2019.

\bibitem{kanerva2009hyperdimensional}
P.~Kanerva, ``Hyperdimensional computing: An introduction to computing in
  distributed representation with high-dimensional random vectors,''
  \emph{Cognitive computation}, vol.~1, no.~2, pp. 139--159, 2009.

\bibitem{rachkovskij2007linear}
D.~Rachkovskij, ``Linear classifiers based on binary distributed
  representations,'' \emph{International Journal Information Theories \&
  Applications}, 2007.

\bibitem{kussul1998application}
E.~M. Kussul, L.~M. Kasatkina, D.~A. Rachkovskij, and D.~C. Wunsch,
  ``Application of random threshold neural networks for diagnostics of micro
  machine tool condition,'' in \emph{1998 IEEE International Joint Conference
  on Neural Networks Proceedings. IEEE World Congress on Computational
  Intelligence (Cat. No. 98CH36227)}, vol.~1.\hskip 1em plus 0.5em minus
  0.4em\relax IEEE, 1998, pp. 241--244.

\bibitem{kussul1991image}
E.~Kussul, ``On image texture recognition by associative-projective
  neurocomputer,'' in \emph{Proceedings of ANNIE'91 Conference, Intelligent
  Engineering Systems through Artificial Neural Networks}.\hskip 1em plus 0.5em
  minus 0.4em\relax ASME Press, 1991, pp. 453--458.

\bibitem{rachkovskij1990audio}
D.~Rachkovskij and T.~Fedoseyeva, ``On audio signals recognition by multilevel
  neural network,'' in \emph{Proceedings of The International Symposium on
  Neural Networks and Neural Computing-NEURONET}, vol.~90, 1990, pp. 281--283.

\bibitem{rahimi2016robust}
A.~Rahimi, P.~Kanerva, and J.~M. Rabaey, ``A robust and energy-efficient
  classifier using brain-inspired hyperdimensional computing,'' in
  \emph{Proceedings of the 2016 International Symposium on Low Power
  Electronics and Design}.\hskip 1em plus 0.5em minus 0.4em\relax ACM, 2016,
  pp. 64--69.

\bibitem{logan2000mel}
B.~Logan \emph{et~al.}, ``Mel frequency cepstral coefficients for music
  modeling.'' in \emph{Ismir}, vol. 270, 2000, pp. 1--11.

\bibitem{rahimi2016hyperdimensional}
A.~Rahimi, S.~Benatti, P.~Kanerva, L.~Benini, and J.~M. Rabaey,
  ``Hyperdimensional biosignal processing: A case study for {EMG}-based hand
  gesture recognition,'' in \emph{2016 IEEE International Conference on
  Rebooting Computing (ICRC)}.\hskip 1em plus 0.5em minus 0.4em\relax IEEE,
  2016, pp. 1--8.

\bibitem{kleyko2014brain}
D.~Kleyko and E.~Osipov, ``Brain-like classifier of temporal patterns,'' in
  \emph{2014 International Conference on Computer and Information Sciences
  (ICCOINS)}.\hskip 1em plus 0.5em minus 0.4em\relax IEEE, 2014, pp. 1--6.

\bibitem{imani2019quanthd}
M.~Imani, S.~Bosch, S.~Datta, S.~Ramakrishna, S.~Salamat, J.~M. Rabaey, and
  T.~Rosing, ``Quanthd: A quantization framework for hyperdimensional
  computing,'' \emph{IEEE Transactions on Computer-Aided Design of Integrated
  Circuits and Systems}, 2019.

\bibitem{rahimi2017hyperdimensional2}
A.~Rahimi, P.~Kanerva, J.~d.~R. Mill{\'a}n, and J.~M. Rabaey,
  ``Hyperdimensional computing for noninvasive brain-computer interfaces: Blind
  and one-shot classification of {EEG} error-related potentials,'' in
  \emph{10th EAI Int. Conf. on Bio-inspired Information and Communications
  Technologies}, no. CONF, 2017.

\bibitem{moin2018emg}
A.~Moin, A.~Zhou, A.~Rahimi, S.~Benatti, A.~Menon, S.~Tamakloe, J.~Ting,
  N.~Yamamoto, Y.~Khan, F.~Burghardt \emph{et~al.}, ``An {EMG} gesture
  recognition system with flexible high-density sensors and brain-inspired
  high-dimensional classifier,'' in \emph{2018 IEEE International Symposium on
  Circuits and Systems (ISCAS)}.\hskip 1em plus 0.5em minus 0.4em\relax IEEE,
  2018, pp. 1--5.

\bibitem{imani2017voicehd}
M.~Imani, D.~Kong, A.~Rahimi, and T.~Rosing, ``Voice{HD}: Hyperdimensional
  computing for efficient speech recognition,'' in \emph{2017 IEEE
  International Conference on Rebooting Computing (ICRC)}.\hskip 1em plus 0.5em
  minus 0.4em\relax IEEE, 2017, pp. 1--8.

\bibitem{imani2018hierarchical}
M.~Imani, C.~Huang, D.~Kong, and T.~Rosing, ``Hierarchical hyperdimensional
  computing for energy efficient classification,'' in \emph{2018 55th
  ACM/ESDA/IEEE Design Automation Conference (DAC)}.\hskip 1em plus 0.5em minus
  0.4em\relax IEEE, 2018, pp. 1--6.

\bibitem{joshi2016language}
A.~Joshi, J.~T. Halseth, and P.~Kanerva, ``Language geometry using random
  indexing,'' in \emph{International Symposium on Quantum Interaction}.\hskip
  1em plus 0.5em minus 0.4em\relax Springer, 2016, pp. 265--274.

\bibitem{imani2018hdna}
M.~Imani, T.~Nassar, A.~Rahimi, and T.~Rosing, ``{HDNA}: Energy-efficient {DNA}
  sequencing using hyperdimensional computing,'' in \emph{2018 IEEE EMBS
  International Conference on Biomedical \& Health Informatics (BHI)}.\hskip
  1em plus 0.5em minus 0.4em\relax IEEE, 2018, pp. 271--274.

\bibitem{kleyko2018classification}
D.~Kleyko, A.~Rahimi, D.~A. Rachkovskij, E.~Osipov, and J.~M. Rabaey,
  ``Classification and recall with binary hyperdimensional computing: Tradeoffs
  in choice of density and mapping characteristics,'' \emph{IEEE transactions
  on neural networks and learning systems}, vol.~29, no.~12, pp. 5880--5898,
  2018.

\bibitem{gayler2004vector}
R.~W. Gayler, ``Vector symbolic architectures answer jackendoff's challenges
  for cognitive neuroscience,'' \emph{arXiv preprint cs/0412059}, 2004.

\bibitem{rahimi2018efficient}
A.~Rahimi, P.~Kanerva, L.~Benini, and J.~M. Rabaey, ``Efficient biosignal
  processing using hyperdimensional computing: Network templates for combined
  learning and classification of {ExG} signals,'' \emph{Proceedings of the
  IEEE}, vol. 107, no.~1, pp. 123--143, 2018.

\bibitem{imani2019binary}
M.~Imani, J.~Messerly, F.~Wu, W.~Pi, and T.~Rosing, ``A binary learning
  framework for hyperdimensional computing,'' in \emph{2019 Design, Automation
  \& Test in Europe Conference \& Exhibition (DATE)}.\hskip 1em plus 0.5em
  minus 0.4em\relax IEEE, 2019, pp. 126--131.

\bibitem{rachkovskij2001representation}
D.~A. Rachkovskij, ``Representation and processing of structures with binary
  sparse distributed codes,'' \emph{IEEE Transactions on Knowledge and Data
  Engineering}, vol.~13, no.~2, pp. 261--276, 2001.

\bibitem{imani2019sparsehd}
M.~Imani, S.~Salamat, B.~Khaleghi, M.~Samragh, F.~Koushanfar, and T.~Rosing,
  ``Sparse{HD}: Algorithm-hardware co-optimization for efficient
  high-dimensional computing,'' in \emph{2019 IEEE 27th Annual International
  Symposium on Field-Programmable Custom Computing Machines (FCCM)}.\hskip 1em
  plus 0.5em minus 0.4em\relax IEEE, 2019, pp. 190--198.

\bibitem{salamat2019f5}
S.~Salamat, M.~Imani, B.~Khaleghi, and T.~Rosing, ``F5-hd: Fast flexible
  {FPGA}-based framework for refreshing hyperdimensional computing,'' in
  \emph{Proceedings of the 2019 ACM/SIGDA International Symposium on
  Field-Programmable Gate Arrays}, 2019, pp. 53--62.

\bibitem{in-mem2019}
\BIBentryALTinterwordspacing
G.~Karunaratne, M.~L. Gallo, G.~Cherubini, L.~Benini, A.~Rahimi, and
  A.~Sebastian, ``In-memory hyperdimensional computing,'' \emph{CoRR}, vol.
  abs/1906.01548, 2019. [Online]. Available:
  \url{http://arxiv.org/abs/1906.01548}
\BIBentrySTDinterwordspacing

\bibitem{rahimi2018hyperdimensional}
A.~Rahimi, T.~F. Wu, H.~Li, J.~M. Rabaey, H.-S.~P. Wong, M.~M. Shulaker, and
  S.~Mitra, ``Hyperdimensional computing nanosystem,'' \emph{arXiv preprint
  arXiv:1811.09557}, 2018.

\bibitem{wu2018hyperdimensional}
T.~F. Wu, H.~Li, P.-C. Huang, A.~Rahimi, G.~Hills, B.~Hodson, W.~Hwang, J.~M.
  Rabaey, H.-S.~P. Wong, M.~M. Shulaker \emph{et~al.}, ``Hyperdimensional
  computing exploiting carbon nanotube {FETs}, resistive {RAM}, and their
  monolithic {3D} integration,'' \emph{IEEE Journal of Solid-State Circuits},
  vol.~53, no.~11, pp. 3183--3196, 2018.

\bibitem{li2016hyperdimensional}
H.~Li, T.~F. Wu, A.~Rahimi, K.-S. Li, M.~Rusch, C.-H. Lin, J.-L. Hsu, M.~M.
  Sabry, S.~B. Eryilmaz, J.~Sohn \emph{et~al.}, ``Hyperdimensional computing
  with {3D VRRAM} in-memory kernels: Device-architecture co-design for
  energy-efficient, error-resilient language recognition,'' in \emph{2016 IEEE
  International Electron Devices Meeting (IEDM)}.\hskip 1em plus 0.5em minus
  0.4em\relax IEEE, 2016, pp. 16--1.

\bibitem{burrello2019laelaps}
A.~Burrello, L.~Cavigelli, K.~Schindler, L.~Benini, and A.~Rahimi, ``Laelaps:
  An energy-efficient seizure detection algorithm from long-term human {iEEG}
  recordings without false alarms,'' in \emph{2019 Design, Automation \& Test
  in Europe Conference \& Exhibition (DATE)}.\hskip 1em plus 0.5em minus
  0.4em\relax IEEE, 2019, pp. 752--757.

\bibitem{isolet}
``{UCI} machine learning repository,''
  \url{http://archive.ics.uci.edu/ml/datasets/ISOLET}.

\bibitem{zhang2014seizure}
Z.~Zhang and K.~K. Parhi, ``Seizure detection using wavelet decomposition of
  the prediction error signal from a single channel of intra-cranial {EEG},''
  in \emph{2014 36th annual international conference of the IEEE engineering in
  medicine and biology society}.\hskip 1em plus 0.5em minus 0.4em\relax IEEE,
  2014, pp. 4443--4446.

\bibitem{zhang2015low}
------, ``Low-complexity seizure prediction from {iEEG/sEEG} using spectral
  power and ratios of spectral power,'' \emph{IEEE transactions on biomedical
  circuits and systems}, vol.~10, no.~3, pp. 693--706, 2015.

\bibitem{zhang2015seizure}
------, ``Seizure detection using regression tree based feature selection and
  polynomial {SVM} classification,'' in \emph{2015 37th Annual International
  Conference of the IEEE Engineering in Medicine and Biology Society
  (EMBC)}.\hskip 1em plus 0.5em minus 0.4em\relax IEEE, 2015, pp. 6578--6581.

\bibitem{zhang2015seizure2}
------, ``Seizure prediction using polynomial {SVM} classification,'' in
  \emph{2015 37th Annual International Conference of the IEEE Engineering in
  Medicine and Biology Society (EMBC)}.\hskip 1em plus 0.5em minus 0.4em\relax
  IEEE, 2015, pp. 5748--5751.

\bibitem{parhi2019discriminative}
K.~K. Parhi and Z.~Zhang, ``Discriminative ratio of spectral power and relative
  power features derived via frequency-domain model ratio with application to
  seizure prediction,'' \emph{IEEE transactions on biomedical circuits and
  systems}, vol.~13, no.~4, pp. 645--657, 2019.

\bibitem{gallant2013representing}
S.~I. Gallant and T.~W. Okaywe, ``Representing objects, relations, and
  sequences,'' \emph{Neural computation}, vol.~25, no.~8, pp. 2038--2078, 2013.

\bibitem{morris2019comphd}
J.~Morris, M.~Imani, S.~Bosch, A.~Thomas, H.~Shu, and T.~Rosing, ``Comphd:
  Efficient hyperdimensional computing using model compression,'' in \emph{2019
  IEEE/ACM International Symposium on Low Power Electronics and Design
  (ISLPED)}.\hskip 1em plus 0.5em minus 0.4em\relax IEEE, 2019, pp. 1--6.

\bibitem{hadamard}
``Hadamard matrix,''
  \url{https://docs.scipy.org/doc/scipy-0.14.0/reference/generated/scipy.linalg.hadamard.html}.

\bibitem{imani2019semihd}
M.~Imani, S.~Bosch, M.~Javaheripi, B.~Rouhani, X.~Wu, F.~Koushanfar, and
  T.~Rosing, ``Semihd: Semi-supervised learning using hyperdimensional
  computing,'' in \emph{IEEE/ACM International Conference On Computer Aided
  Design (ICCAD)}, 2019, pp. 1--8.

\bibitem{imani2019adapthd}
M.~Imani, J.~Morris, S.~Bosch, H.~Shu, G.~De~Micheli, and T.~Rosing, ``Adapthd:
  Adaptive efficient training for brain-inspired hyperdimensional computing,''
  in \emph{2019 IEEE Biomedical Circuits and Systems Conference
  (BioCAS)}.\hskip 1em plus 0.5em minus 0.4em\relax IEEE, 2019, pp. 1--4.

\bibitem{haykin2014adaptive}
S.~S. Haykin, \emph{Adaptive filter theory}.\hskip 1em plus 0.5em minus
  0.4em\relax Pearson Education India, 2014.

\bibitem{manabat2019performance}
A.~X. Manabat, C.~R. Marcelo, A.~L. Quinquito, and A.~Alvarez, ``Performance
  analysis of hyperdimensional computing for character recognition,'' in
  \emph{2019 International Symposium on Multimedia and Communication Technology
  (ISMAC)}.\hskip 1em plus 0.5em minus 0.4em\relax IEEE, 2019, pp. 1--5.

\bibitem{kanerva2000random}
P.~Kanerva, J.~Kristoferson, and A.~Holst, ``Random indexing of text samples
  for latent semantic analysis,'' in \emph{Proceedings of the Annual Meeting of
  the Cognitive Science Society}, vol.~22, no.~22, 2000.

\bibitem{recchia2015encoding}
G.~Recchia, M.~Sahlgren, P.~Kanerva, and M.~N. Jones, ``Encoding sequential
  information in semantic space models: Comparing holographic reduced
  representation and random permutation,'' \emph{Computational intelligence and
  neuroscience}, vol. 2015, 2015.

\bibitem{kleyko2019distributed}
D.~Kleyko, E.~Osipov, D.~De~Silva, U.~Wiklund, V.~Vyatkin, and D.~Alahakoon,
  ``Distributed representation of n-gram statistics for boosting
  self-organizing maps with hyperdimensional computing,'' in
  \emph{International Andrei Ershov Memorial Conference on Perspectives of
  System Informatics}.\hskip 1em plus 0.5em minus 0.4em\relax Springer, 2019,
  pp. 64--79.

\bibitem{bandaragoda2019trajectory}
T.~Bandaragoda, D.~De~Silva, D.~Kleyko, E.~Osipov, U.~Wiklund, and
  D.~Alahakoon, ``Trajectory clustering of road traffic in urban environments
  using incremental machine learning in combination with hyperdimensional
  computing,'' in \emph{2019 IEEE Intelligent Transportation Systems Conference
  (ITSC)}.\hskip 1em plus 0.5em minus 0.4em\relax IEEE, 2019, pp. 1664--1670.

\bibitem{hersche2020evolvable}
M.~Hersche, S.~Sangalli, L.~Benini, and A.~Rahimi, ``Evolvable hyperdimensional
  computing: Unsupervised regeneration of associative memory to recover faulty
  components,'' in \emph{IEEE International Conference on Artificial
  Intelligence Circuits and Systems (AICAS), Genoa, Italy, March 23-25,
  2020}.\hskip 1em plus 0.5em minus 0.4em\relax IEEE, 2020.

\bibitem{kleyko2016holographic}
D.~Kleyko, E.~Osipov, A.~Senior, A.~I. Khan, and Y.~A.
  {\c{S}}ekerciog{\u{g}}lu, ``Holographic graph neuron: A bioinspired
  architecture for pattern processing,'' \emph{IEEE transactions on neural
  networks and learning systems}, vol.~28, no.~6, pp. 1250--1262, 2016.

\bibitem{triguero2015self}
I.~Triguero, S.~Garc{\'\i}a, and F.~Herrera, ``Self-labeled techniques for
  semi-supervised learning: taxonomy, software and empirical study,''
  \emph{Knowledge and Information systems}, vol.~42, no.~2, pp. 245--284, 2015.

\bibitem{najafabadi2016hyperdimensional}
F.~R. Najafabadi, A.~Rahimi, P.~Kanerva, and J.~M. Rabaey, ``Hyperdimensional
  computing for text classification,'' in \emph{Design, Automation Test in
  Europe Conference Exhibition (DATE), University Booth}, 2016, pp. 1--1.

\bibitem{rahimi2017hyperdimensional}
A.~Rahimi, A.~Tchouprina, P.~Kanerva, J.~d.~R. Mill{\'a}n, and J.~M. Rabaey,
  ``Hyperdimensional computing for blind and one-shot classification of {EEG}
  error-related potentials,'' \emph{Mobile Networks and Applications}, pp.
  1--12, 2017.

\bibitem{burrello2018one}
A.~Burrello, K.~Schindler, L.~Benini, and A.~Rahimi, ``One-shot learning for
  {iEEG} seizure detection using end-to-end binary operations: Local binary
  patterns with hyperdimensional computing,'' in \emph{2018 IEEE Biomedical
  Circuits and Systems Conference (BioCAS)}.\hskip 1em plus 0.5em minus
  0.4em\relax IEEE, 2018, pp. 1--4.

\bibitem{rasanen2014modeling}
O.~R{\"a}s{\"a}nen and S.~Kakouros, ``Modeling dependencies in multiple
  parallel data streams with hyperdimensional computing,'' \emph{IEEE Signal
  Processing Letters}, vol.~21, no.~7, pp. 899--903, 2014.

\bibitem{rasanen2015sequence}
O.~J. R{\"a}s{\"a}nen and J.~P. Saarinen, ``Sequence prediction with sparse
  distributed hyperdimensional coding applied to the analysis of mobile phone
  use patterns,'' \emph{IEEE transactions on neural networks and learning
  systems}, vol.~27, no.~9, pp. 1878--1889, 2015.

\bibitem{kleyko2015imitation}
D.~Kleyko, E.~Osipov, R.~W. Gayler, A.~I. Khan, and A.~G. Dyer, ``Imitation of
  honey bees’ concept learning processes using vector symbolic
  architectures,'' \emph{Biologically Inspired Cognitive Architectures},
  vol.~14, pp. 57--72, 2015.

\bibitem{yilmaz2015connectionist}
O.~Yilmaz, ``Connectionist-symbolic machine intelligence using cellular
  automata based reservoir-hyperdimensional computing,'' \emph{arXiv preprint
  arXiv:1503.00851}, 2015.

\bibitem{kleyko2017modality}
D.~Kleyko, S.~Khan, E.~Osipov, and S.-P. Yong, ``Modality classification of
  medical images with distributed representations based on cellular automata
  reservoir computing,'' in \emph{2017 IEEE 14th International Symposium on
  Biomedical Imaging (ISBI 2017)}.\hskip 1em plus 0.5em minus 0.4em\relax IEEE,
  2017, pp. 1053--1056.

\bibitem{montone2017hyper}
G.~Montone, J.~K. O'Regan, and A.~V. Terekhov, ``Hyper-dimensional computing
  for a visual question-answering system that is trainable end-to-end,''
  \emph{arXiv preprint arXiv:1711.10185}, 2017.

\bibitem{kleyko2018hyperdimensional}
D.~Kleyko, E.~Osipov, N.~Papakonstantinou, and V.~Vyatkin, ``Hyperdimensional
  computing in industrial systems: the use-case of distributed fault isolation
  in a power plant,'' \emph{IEEE Access}, vol.~6, pp. 30\,766--30\,777, 2018.

\bibitem{mitrokhin2019learning}
A.~Mitrokhin, P.~Sutor, C.~Ferm{\"u}ller, and Y.~Aloimonos, ``Learning
  sensorimotor control with neuromorphic sensors: Toward hyperdimensional
  active perception,'' \emph{Science Robotics}, vol.~4, no.~30, p. eaaw6736,
  2019.

\bibitem{rasanen2015generating}
O.~J. R{\"a}s{\"a}nen, ``Generating hyperdimensional distributed
  representations from continuous-valued multivariate sensory input.'' in
  \emph{CogSci}, 2015.

\bibitem{bosch2019qubithd}
S.~Bosch, A.~S. de~la Cerda, M.~Imani, T.~S. Rosing, and G.~De~Micheli,
  ``{QubitHD}: A stochastic acceleration method for {HD} computing-based
  machine learning,'' \emph{arXiv preprint arXiv:1911.12446}, 2019.

\bibitem{imani2019framework}
M.~Imani, Y.~Kim, S.~Riazi, J.~Messerly, P.~Liu, F.~Koushanfar, and T.~Rosing,
  ``A framework for collaborative learning in secure high-dimensional space,''
  in \emph{2019 IEEE 12th International Conference on Cloud Computing
  (CLOUD)}.\hskip 1em plus 0.5em minus 0.4em\relax IEEE, 2019, pp. 435--446.

\bibitem{zhang2018muse}
Z.~Zhang and K.~K. Parhi, ``{MUSE}: Minimum uncertainty and sample elimination
  based binary feature selection,'' \emph{IEEE Transactions on Knowledge and
  Data Engineering}, vol.~31, no.~9, pp. 1750--1764, 2018.

\bibitem{alonso2020hyperembed}
P.~Alonso, K.~Shridhar, D.~Kleyko, E.~Osipov, and M.~Liwicki, ``{HyperEmbed}:
  Tradeoffs between resources and performance in {NLP} tasks with
  hyperdimensional computing enabled embedding of n-gram statistics,''
  \emph{arXiv preprint arXiv:2003.01821}, 2020.

\bibitem{kleyko2019density}
D.~Kleyko, M.~Kheffache, E.~P. Frady, U.~Wiklund, and E.~Osipov, ``Density
  encoding enables resource-efficient randomly connected neural networks,''
  \emph{arXiv preprint arXiv:1909.09153}, 2019.

\bibitem{kleyko2017integer}
D.~Kleyko, E.~P. Frady, and E.~Osipov, ``Integer echo state networks:
  {H}yperdimensional reservoir computing,'' \emph{arXiv preprint
  arXiv:1706.00280}, 2017.

\bibitem{anderson2017high}
A.~G. Anderson and C.~P. Berg, ``The high-dimensional geometry of binary neural
  networks,'' \emph{arXiv preprint arXiv:1705.07199}, 2017.

\end{thebibliography}

\end{document}